\documentclass[runningheads]{llncs}

\usepackage{amsmath}
\usepackage{amssymb}
\usepackage{bbm}
\usepackage{booktabs} 
\usepackage{color-edits}
\usepackage{color}
\usepackage{comment}
\usepackage{epsfig}
\usepackage{graphicx}
\usepackage{hyperref}
\usepackage{microtype}
\usepackage{multirow}
\usepackage{paralist}
\usepackage{subfigure}
\usepackage{sidecap}
\usepackage{url}
\usepackage[dvipsnames]{xcolor}
\usepackage{wrapfig}

\addauthor{pg}{red}
\addauthor{sg}{green}
\addauthor{ss}{orange}

\begin{document}

\pagestyle{headings}
\mainmatter
\def\ECCVSubNumber{7019}  

\title{Cross-modal Learning for Multi-modal \\ Video Categorization} 

\author{Palash Goyal\thanks{Equal contribution} \and
Saurabh Sahu* \and
Shalini Ghosh* \and
Chul Lee}
\institute{VDI Lab, Samsung Research America, USA
}
\authorrunning{ }
\titlerunning{}

\maketitle

\begin{abstract}
Multi-modal machine learning (ML) models can process data in multiple modalities (e.g., video, audio, text) and are useful for video content analysis in a variety of problems (e.g., object detection, scene understanding, activity recognition). In this paper, we focus on the problem of video categorization using a multi-modal ML technique. In particular, we have developed a novel multi-modal ML approach that we call ``cross-modal learning'', where one modality influences another but only when there is correlation between the modalities --- for that, we first train a correlation tower that guides the main multi-modal video categorization tower in the model. We show how this cross-modal principle can be applied to different types of models (e.g., RNN, Transformer, NetVLAD), and demonstrate through experiments how our proposed multi-modal video categorization models with cross-modal learning out-perform strong state-of-the-art baseline models.
\end{abstract}

\section{Introduction}
\label{sec:intro}

Multi-modal machine learning (ML) models typically process data from different modalities (e.g., video, image, audio, language) in order to perform a task. For example, consider the task of identifying categories in a video. A ML model may be able to detect objects like {\tt guitar} and {\tt crowd} in the video frames --- when this is combined with the audio channel inferring {\tt rock music}, it could help the ML model infer with a high confidence that the video category is {\tt rock concert}. 

Multi-modal ML models have some advantages: (a) higher accuracy, since data from different modalities can reinforce each other or help in disambiguation; (b) resilience to errors in the data, e.g., if the video channel goes dark due to a data glitch, close-caption text of relevant words combined with audio of the sound of gunfire and explosions could help detect that the video segment is an action sequence from a war movie. These advantages can make multi-modal ML models very suitable for different types of video content analysis  problems. 

A potential area of application of such multi-modal technology is in the content recommendation problem, where the ML model suggests new video content to a user based on whether characteristics in that video matches the corresponding characteristics inferred from the sequence of watched videos in the user's history.
The characteristics of the video that we can consider in this context are fine-grained categories (e.g., Entertainment/Concert, Action/War).

A key innovation that we propose in this paper is {\em cross-modal learning}, where we use the information from one modality to positively influence the others when doing a fusion of different modalities in the model. There are different ways of doing multi-modal fusion in an ML model: (a) In ``early fusion", the input embeddings from the different modalities are concatenated and then usually passed through a ReLU layer to allow cross-modal interaction between the input embeddings --- this essentially learns a joint input embedding between the different modalities, by considering a product between the input embedding spaces of the different modalities; (b) On the other hand, in ``late fusion" the output embeddings of the towers corresponding to different modalities are usually combined through a ReLU --- this learns a joint output embedding between the different modalities; (c) In ``intermediate fusion", the cross-modal combination happens in an intermediate stage --- this is ideal for incorporating domain knowledge of various forms from one modality to another, e.g., let the attention vectors from one modality influence the other, especially if the modalities are correlated with each other.
Cross-modal fusion is a type of {\em intermediate fusion}.

Our key contributions in this paper are:
\begin{enumerate}
\item Proposing a novel mechanism of intermediate fusion in multi-modal learning that we call ``cross-modal learning'', where the intermediate output from one modality (e.g., attention vectors in RNN model) are used to influence the model components corresponding to other modalities in the data.
\item Improving the cross-modal learning mechanism by using correlation towers in the model, which identifies when one modality is correlated with other modalities in the data and controls the use of cross-modal learning accordingly.
\item Showing the mathematical formulation of how the principle of cross-modal fusion can be incorporated into different types of models, specifically attention RNN~\cite{Bahdanau14}, Transformer~\cite{Vaswani17} and 
NetVLAD~\cite{Arandjelovic16}.
\item  Creating a hierarchical taxonomy of labels spanning 640 labels for incorporating more semantics and mapping each YouTube-8M category to our taxonomy (with the help of raters).
\item Running large-scale experiments on the YouTube-8M data with our hierarchical taxonomy, where we show how cross-modal learning with correlation tower gives significant improvements for video category prediction over a state-of-the-art baseline.
\item Our experiments with cross-modal learning show that using the multi-modal correlation prediction as a (concatenated) feature rather than in a gating logic gives higher gains, and certain categories in the data show more gains with cross-modal learning than others.
\end{enumerate}

Section~\ref{sec:cross-modal} discusses the main formulation of cross-modal learning and how it can be applied to different models, e.g., attention RNN, Transformer and NetVLAD. Section~\ref{sec:correlation} outlines how cross-modal learning can be further improved by using correlation between the modalities, learned separately using ``correlation towers". Section~\ref{sec:experiments} discusses the experimental setup and Section~\ref{sec:results} analyzes the results, especially showing how using cross-modal learning gives state-of-the-art performance. Section~\ref{sec:related}  outlines related research, while Section~\ref{sec:conclusions} concludes the paper and discusses possible future work.

\section{Cross-modal Learning}
\label{sec:cross-modal}

In multi-modal learning, one important issue to consider is that the modalities are often not completely synchronized with each other in a particular video. For instance, the features detecting a category in the video and the audio can be out of sync --- in a rock concert video, we may hear the rock music in the audio track before the video starts showing salient objects like guitar, crowds. In such cases, the attention for the video may be related to the audio but only after some sort of transform is applied, e.g., a shift or compression of the attention vector in one modality is a good prior for the other modality.
\vspace{-8mm}
\begin{figure}[!hbtp]
\centering
\begin{tabular}
{@{\hspace{-0.2cm}}c@{\hspace{0.0cm}}}
{\includegraphics[width=0.5\textwidth]{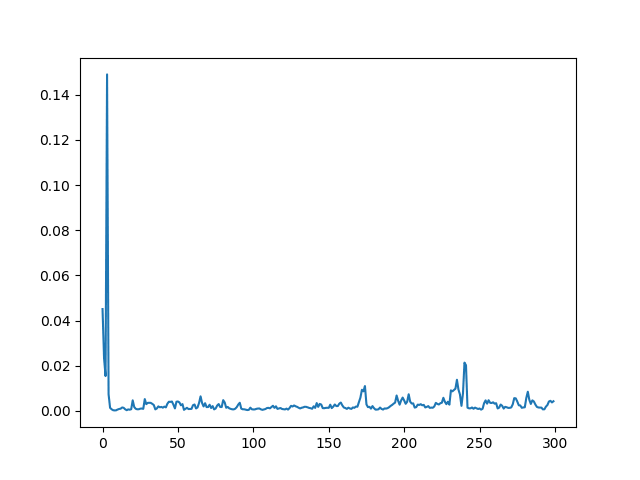}} 
{\includegraphics[width=0.5\textwidth]{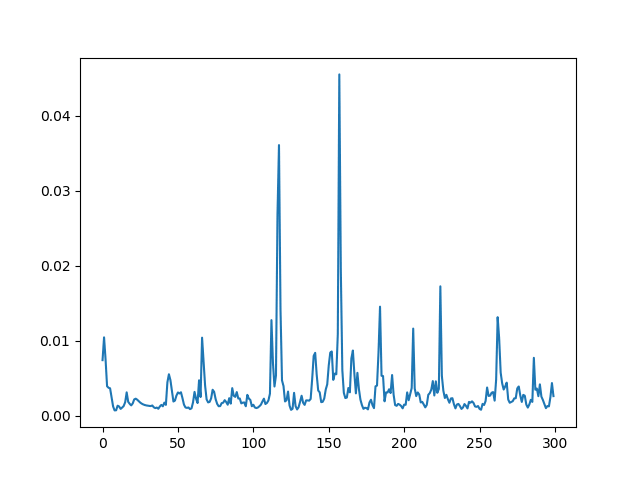}}
\end{tabular}
\vspace{-4mm}
\caption{Attention weights for video (left) and audio (right) modalities for a video.}
\label{fig:attn}
\vspace{-6mm}
\end{figure}

We notice this in our experimental evaluation also. Figure~\ref{fig:attn} shows the attention distributions for an attentional RNN model corresponding to a particular input video --- we see that the audio channel attention is more distributed, while the video channel attention is more peaked and shifted w.r.t. the audio attention. This observation motivated us to consider that when we use the attention of one modality to modulate the attention of another modality, we should consider the attention vector to be transformed via scaling and linear shift, i.e., via a ReLU transform.
In this section, we outline in detail how such cross-modal learning with suitable transforms is incorporated into different ML models, specifically attention RNN,  Transformer and NetVLAD. \\

\begin{figure}[hbtp]
\centering
\includegraphics[width=0.9\textwidth]{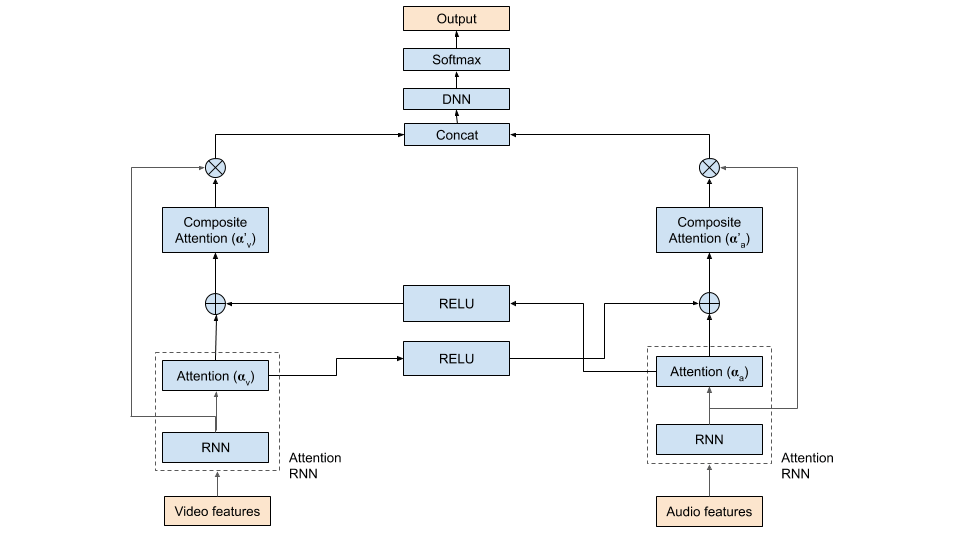}
\caption{RNN Model~\cite{Bahdanau14} with cross-modal attention.}
\label{fig:rnn-cross}
\end{figure}

{\bf RNN:} The equations for the cross-modal attention RNN~\cite{Bahdanau14} update are as follows (Figure~\ref{fig:rnn-cross}):
\begin{eqnarray*}
c_a = \sum 
(\alpha_a + F_v(\alpha_v)).h_a \\
c_v = \sum 
(\alpha_v + F_a(\alpha_a)).h_v,
\end{eqnarray*}

where $F_v$ and $F_a$ are linear (ReLU) transform functions, $\alpha_v$ and $\alpha_a$ are the attention vectors (for video and audio respectively),  $c_v$ and $c_a$ are the context vectors  (for video and audio respectively).


When performing our experiments, we noticed that the multi-modal attention cluster formulation of Long et al.~\cite{long2018} gave improved results --- so, the version of cross-modal attention RNN that we consider in our experiments uses the multi-modal attention cluster formulation similar to Long et al. (with one difference --- we do not use the shifting operation). \\

\begin{figure}[hbtp]
\centering
\includegraphics[width=0.9\textwidth]{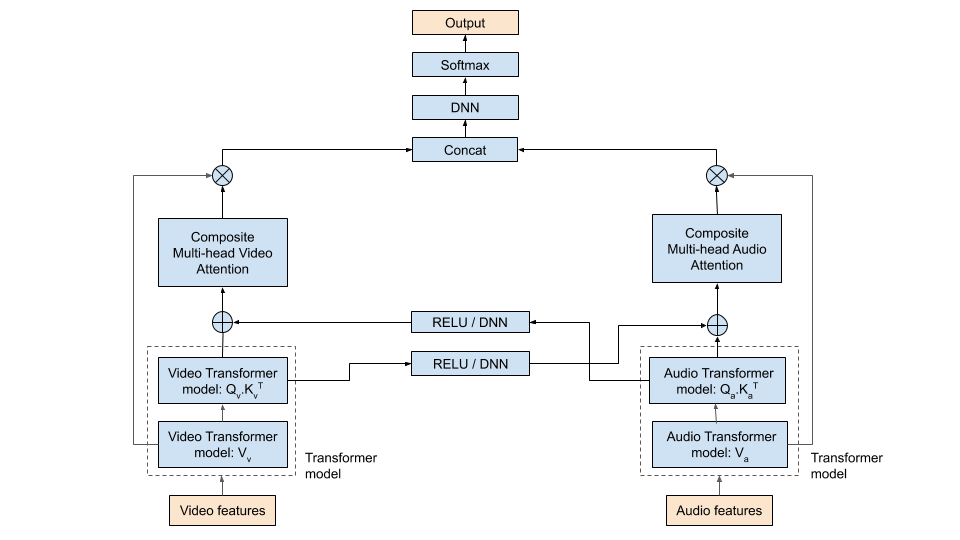}
\caption{Transformer model~\cite{Vaswani17} with cross-modal attention.}
\label{fig:transformer-cross}
\end{figure}

{\bf Transformer:} The equations for the cross-modal Transformer~\cite{Vaswani17} update for the video channel ($v$) are as follows (Figure~\ref{fig:transformer-cross}):
\begin{eqnarray*}
Multihead_v(Q, K, V) = Concat_v(h_1,\ldots,h_k).W_v^O, \\ 
\mbox{where }
h_i = softmax(Q_{iv}W^Q_{iv}  (K_{iv}W^K_{iv})^T + &&F_{ia}(Q_{ia}W^Q_{ia}(K_{ia}W^K_{iv})^T)).VW^V_{iv},
\end{eqnarray*}

where $F_{ia}$ is the ReLU / DNN transform function for the $i^{th}$ projection of the audio features, and
the $W$ matrices are the projection parameter matrices. We have similar update equations for the audio ($a$) channel.\\

\begin{figure}[hbtp]
\centering
\includegraphics[width=0.9\textwidth]{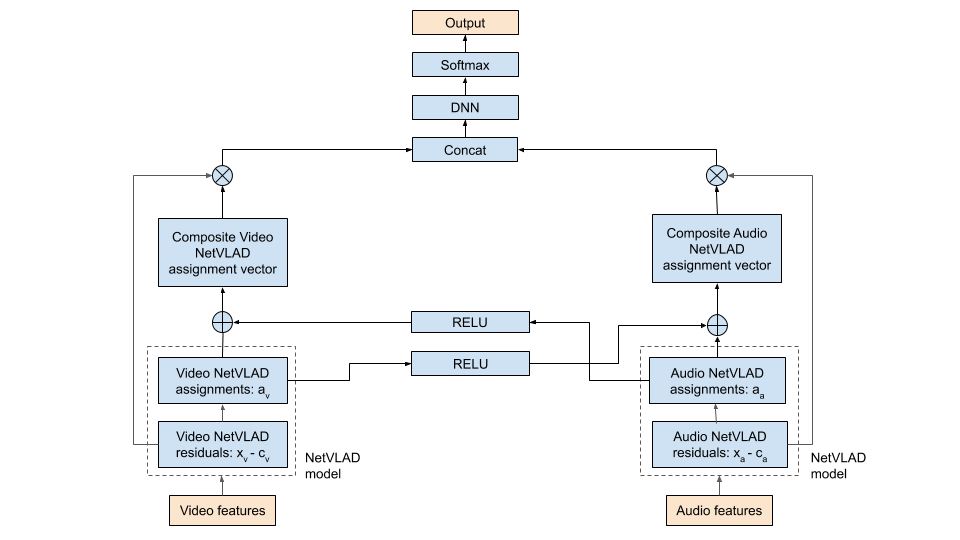}
\caption{NetVLAD model~\cite{Arandjelovic16} with cross-modal attention.}
\label{fig:netvlad-cross}
\end{figure}

{\bf NetVLAD:} The equations for the cross-modal NetVLAD~\cite{Arandjelovic16} update are as follows (Figure~\ref{fig:netvlad-cross}):
\begin{eqnarray*}
V_a = \sum 
(\alpha_a + F_v(\alpha_v)).(x_a - c_a) \\
V_v = \sum 
(\alpha_v + F_a(\alpha_a)).(x_v - c_v),
\end{eqnarray*}

where $F_{\{a,v\}}$ are linear (ReLU) transform functions, $\alpha$ is the assignment vector, $V_{\{a,v\}}$ are the VLAD vectors, $x_{\{a,v\}}-c_{\{a,v\}}$ are the residuals (for both the audio channel $a$ and video channel $v$).\\


\section{Correlation Tower}
\label{sec:correlation}

In some cases the video may be irrelevant with respect to the audio, e.g., the video may shown a static album cover picture while the audio plays the sound. It is also possible for the audio to be irrelevant w.r.t. the video, e.g., home videos often show scenes like dogs playing in a park, while the audio track plays some muzak background music not relevant to the video. 
In these situations, there is no direct correlation between the video and the audio channels --- so, in such cases we do not want the modalities to influence each other.

\begin{figure}[hbtp]
\centering
    \includegraphics[width=0.9\columnwidth]{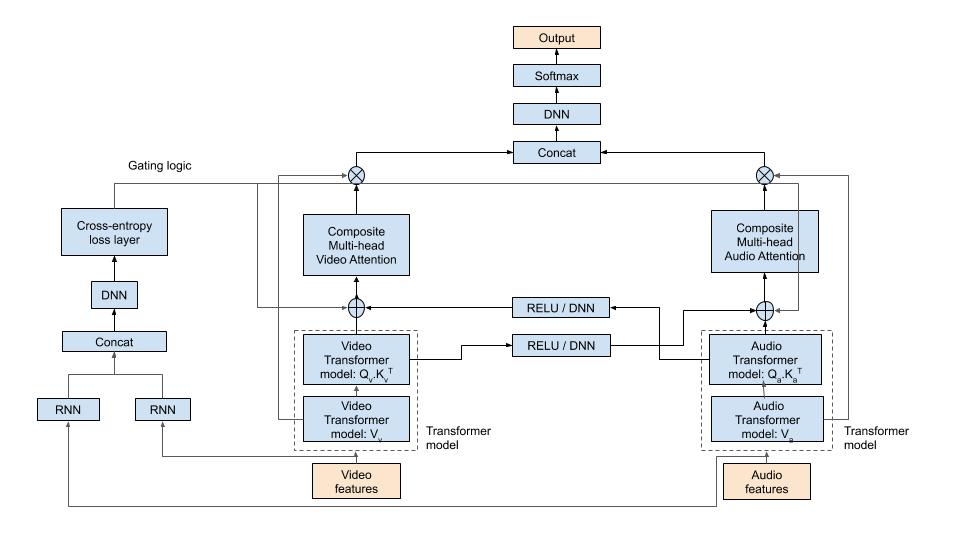}
    \caption{Cross-modal Transformer with cross-entropy based correlation tower.}
    \label{fig:cm-transformer-cross}
\end{figure}

To this end, we train a separate tower on positive and negative examples of audio/video correlation, using the cross-entropy --- Figure~\ref{fig:cm-transformer-cross} shows the relevant architecture for Transformer models (similar architecture is used for RNN and NetVLAD models). The cross entropy loss formulation we use is the following:
\begin{equation}
y\log{y'(f_v, f_a)} + (1-y)\log{(1-y'(f_v, f_a))},
\end{equation}

where $y$ is the indicator of whether an example had positive correlation between the audio and video modalities, $y'$ is the normalized output of the model that takes as input the the feature embeddings of the two modalities ($f_v$ for video and $f_a$ for audio) and predicts whether they are correlated or not. 

The correlation tower is separately learned and the parameters frozen --- the output of the model is used as the gating logic for the cross-modal model. During inference, the output of the cross-entropy loss layer is the prediction of whether the audio and video channels are correlated with each other. This signal is used as a gating logic for the cross-modal attention --- if the prediction is 1, then the modalities are correlated and the cross-modal influence is activated; on the other hand if the prediction is 0, then the modalities are not correlated and the cross-modal influence is de-activated.

One interesting variant of the correlation tower that we considered in our experiments is where the output of one modality was passed through a series of DNNs (say, 2-3 fully connected DNN layers) and then the output was not used to gate the other modality --- instead, the output of the transform was concatenated with the other modality input. This method tended to outperform the gating logic in general --- in hindsight this result seems intuitive, since the concatenation gives the model more freedom to combine the correlation signal with the modalities, whereas gating is a more restrictive logic.

\section{Experiments}
\label{sec:experiments}

In this section, we explain our experimental setting. We first present details of the YouTube-8M data set used for our experiments and the hierarchical taxonomy we created for incorporating more semantics. We then explain the experimental setup specifying the training criteria and hyperparameter selection.

\subsection{Dataset} 
\label{sec:data}


We perform our experiments on the YouTube-8M dataset consisting of 5 million YouTube videos. The videos in the data set are assigned tags from a pre-defined set of ~3800 labels constructed from analyzing text consisting of video title, description and user tags. Each video consists of video and audio signals sampled at 1 frame per second for the first 300 seconds of the video. The frames are passed through Inception-v3 to get a fixed length representation. This was followed by dimensionality reduction using principal component analysis to obtain a 1024 dimensional video and 128 dimensional audio representation. All our experiments use these embeddings as input.

\begin{figure}[hbtp]
    \centering
    \includegraphics[width=0.65\textwidth]{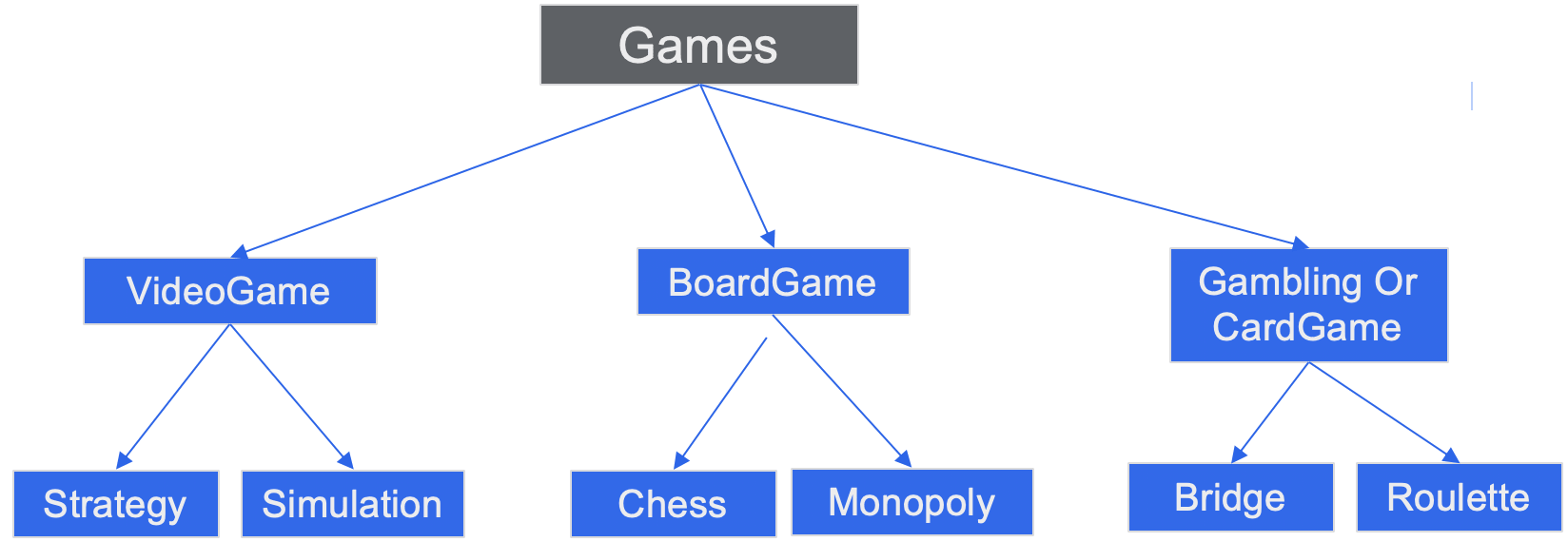}
    \caption{Hierarchical taxonomy for top label ``Games''.}
    \label{fig:hierarchy}
\end{figure}
 
\begin{figure}[hbtp]
    \centering
    \includegraphics[width=0.65\textwidth]{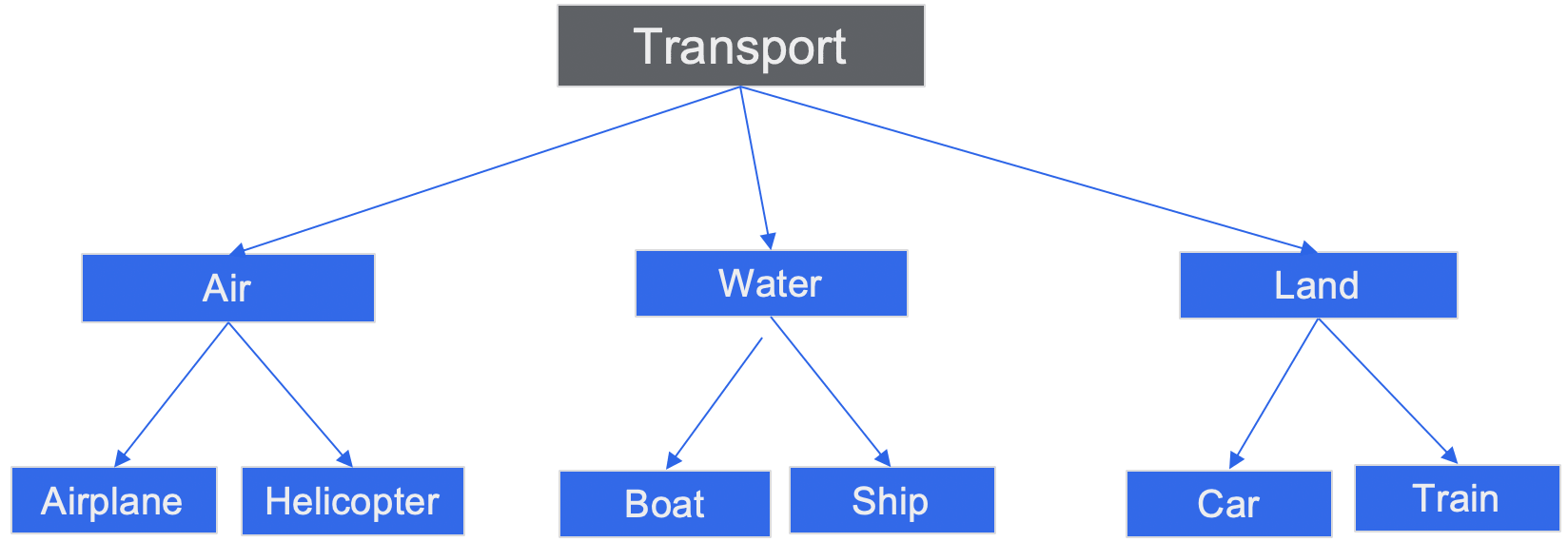}
    \caption{Hierarchical taxonomy for the top label ``Transport''.}
    \label{fig:hierarchy2}
\end{figure}

YouTube-8M video tags do not capture the relation between the labels. For example, tags like ``GTA 5'' and ``GTA 5 (game)'' are semantically same but for loss calculation mutually exclusive. To incorporate semantics, we created a hierarchical taxonomy of labels spanning 640 labels as shown in Figure~\ref{fig:hierarchy} and~\ref{fig:hierarchy2}. 
We then mapped each YouTube-8M category to our taxonomy. The mapping was done by 5-7 annotators and cleaned by 3 engineers.

To train the correlation tower (see Section~\ref{sec:correlation}), we consider visual and audio features from the same video as positive examples for our correlation tower. For negative examples, we consider two videos from different top-level labels from our taxonomy and get visual feature from one and audio feature from the other. 

\subsection{Experimental Setup and Hyperparameters} \label{sec:expt}
We performed all our experiments on a 64 bit Ubuntu 18.0 LTS system with Intel (R) Core (TM) i9-9820X CPU with 10 CPU cores, 3.30 GHz CPU clock speed, 64 GB RAM, and 2 Nvidia GeForce RTX 2080 Ti with 12 GB memory.

For the task of multi-label video categorization, we evaluate our model using four metrics: (i) Global Average Precision, which computes the area under curve of the overall precision-recall curve, (ii) Mean Average Precision, which computes a per class area under curve of precision recall curve and averages them, (iii) Precision at Equal Recall Rate (PERR), which computes the mean precision up to the number of ground truth labels in each class, and (iv) Hit@1, which is the fraction of the test samples that contain a ground truth label in the topmost prediction.

For all the experiments, we train our model on 4 million videos, use 64k videos to tune model hyperparameters and early stopping, and perform 5 test runs with 128k videos each, randomly sampled from the remain 934k videos. We report the mean of the 5 test runs --- standard deviation in each case was less than 0.03.

For comparison, we run state-of-the-art Transformer and NetVLAD models as baselines. We ran a grid search on the Transformer and NetVLAD architectures to identify the best performing architecture and used it for our experiments. We used the same architecture for our variations of Transformer and NetVLAD models as well for fair comparison. Our model has new cross-modal layers. We do a grid search on their sizes and numbers on validation set. We observed that increasing the depth beyond two layers didn't improve in performance. We used Adam optimizer for training with initial learning rate of 0.0002, batch size of 128 and a scheduler for decreasing the learning rate on validation set with patience 2. We padded 0 frames at the end of each video to get 300 frames per video. Each model was trained for a maximum of 10 epochs.

To showcase the performance of our cross modal layer and correlation tower, we report results on the Transformer model in the rest of this section. As shown in the above sections, the approach can be generalized to other models such as RNN as well.

\section{Results} 
\label{sec:results}

In this section, we illustrate the improvement in performance of video categorization by using cross-modal learning and correlation tower. We first analyze the correlation tower by itself and then show results when we use its output to enhance cross-modal learning. For both Transformer and NetVLAD, we show quantitative numbers illustrating the improvement in performance, qualitative results to show examples for which the our cross-modal layer helps in categorization and finally, error analysis where we show examples where our model makes faulty predictions, and discuss how we can further mitigate them.


\begin{figure}[hb!]
    \centering
        \includegraphics[width=0.7\textwidth]{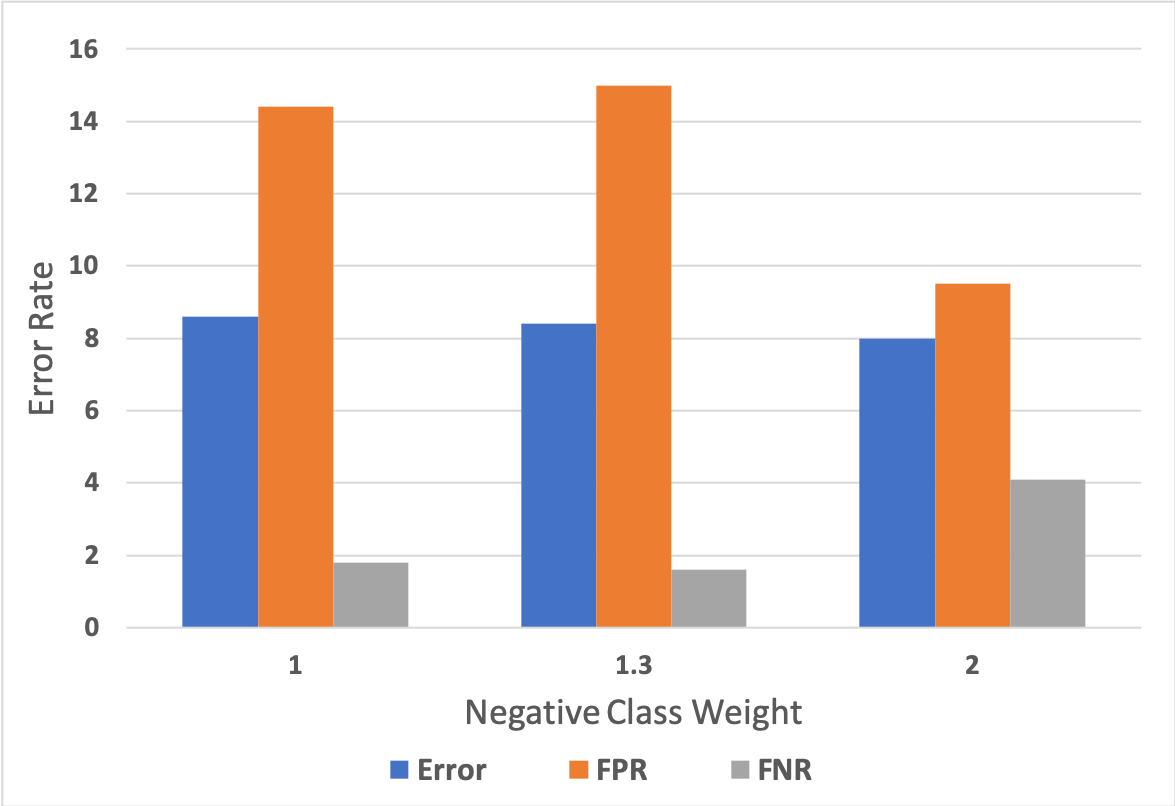}
    \caption{Performance of audio-video synchronization tower with varying negative class weight. FPR and FNR stand for false positive rate and false negative rate respectively.}
    \label{fig:synTower1}
\end{figure}
\begin{figure}[hb!]
        \centering
        \includegraphics[width=0.7\textwidth]{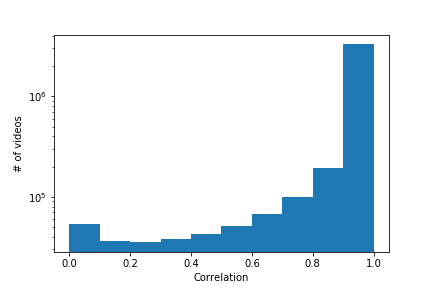}
    \caption{Distribution of correlation tower output on training data (negative weight 2).}
    \label{fig:synTower2}
\end{figure}

\subsection{Correlation Tower}
\subsubsection{Quantitative Results}

We present the results of the correlation tower performance in Figure~\ref{fig:synTower1} --- it shows overall error rate, false positive rate (FPR) and false negative rate (FNR). In our data set, taking video and audio from different categories ensures dissimilarity. However, a given video may have (i) no audio data, (ii) irrelevant audio (many examples have music), or (iii) noisy audio. Further, the goal of our correlation tower is to identify uncorrelated visual and audio features from a video. To address this, we give higher weight to negative examples.
We observe that it reduces overall error rate and also gives a good trade-off between false positive rate and false negative rate. Overall, the correlation tower gives an accuracy of 92\%, and FPR and FNR of 9.6\% and 4.1\% respectively.  
We also show the distribution of output predictions for positive classes correlation tower in Figure~\ref{fig:synTower2} --- we observe that although over 90\% examples are identified as correlated, there are several videos with low correlations. We further analyze the false positive and false negative examples now showing that many of these supposed false negatives are actually true negatives. 

\subsubsection{Qualitative Results}
\label{sec:corrTower}

\begin{table}[hbtp]
\vspace{-10mm}
\centering
\caption{Qualitative analysis of correlation tower. For each of the examples below, we feed visual and audio features from the same video, but the correlation tower predicted the audio and the video features for each of the examples to be uncorrelated, as desired.}
\label{tab:corr_qual1}
\scriptsize
\begin{tabular}{l|c|c}
 FN           &    Example 1      &  Example 2  \\ \hline \hline
Video       &    \includegraphics[width=0.35\columnwidth]{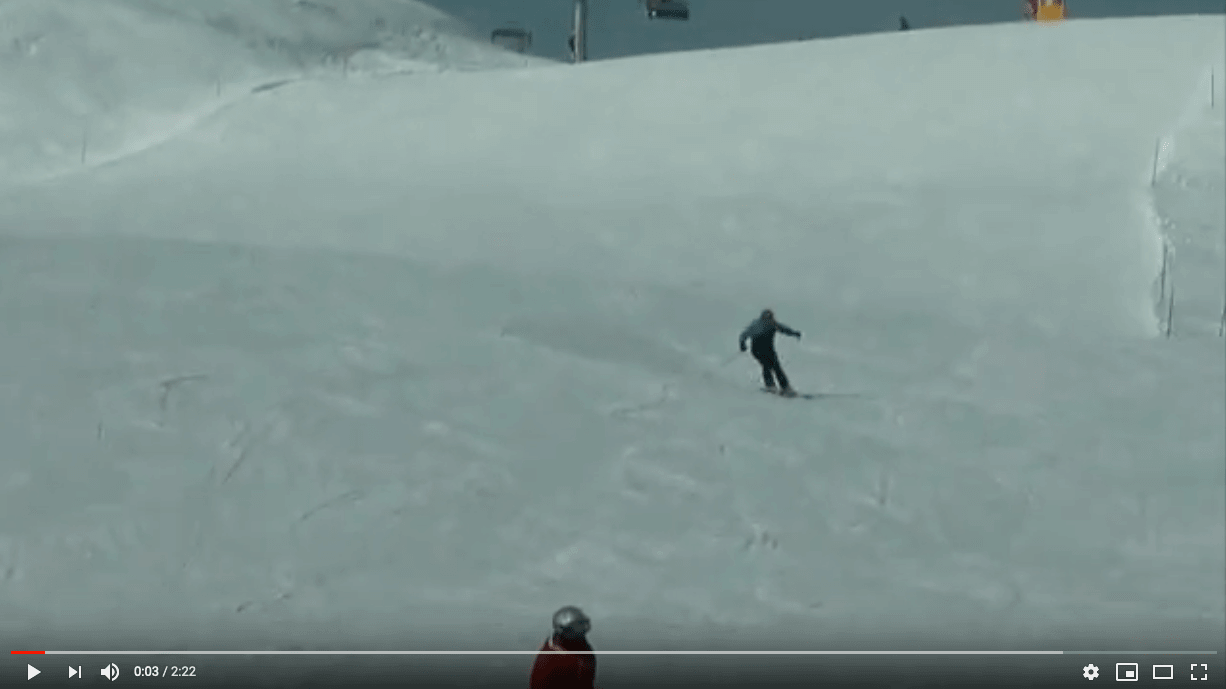}   &  \includegraphics[width=0.35\columnwidth]{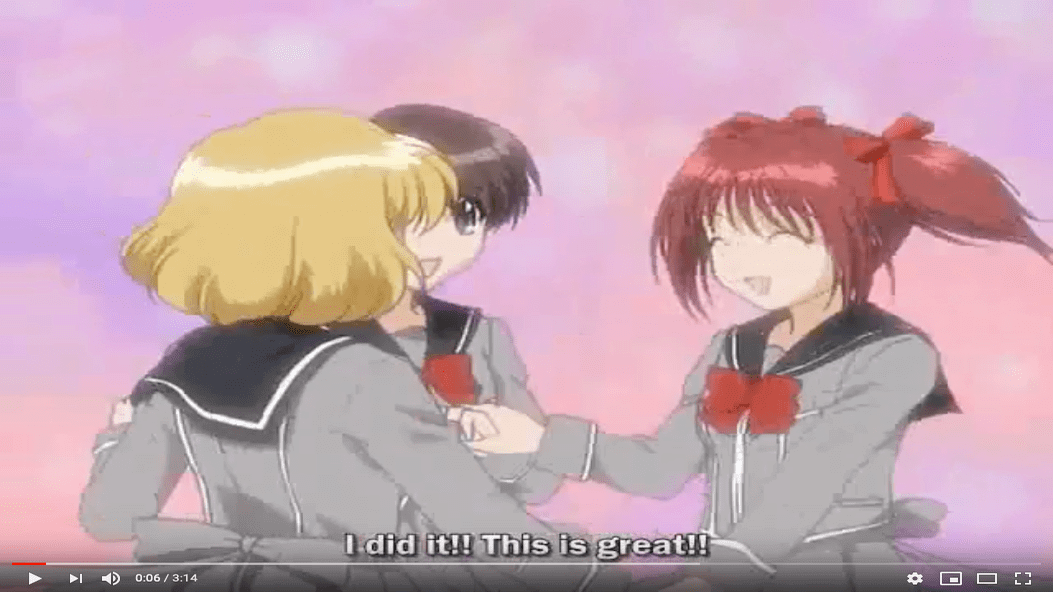}          \\ \hline\hline
Link & bit.ly/39hxAtM & bit.ly/39nKS7Y \\ \hline \hline
Content(v) & Skiing & Anime \\ \hline \hline
Content(a) & Music & Music \\ \hline \hline
Prediction & Uncorrelated & Uncorrelated \\ \hline \hline
\end{tabular}
\vspace{-6mm}
\end{table}

\begin{table}[hbtp]
\vspace{-3mm}
\centering
\caption{Qualitative analysis of correlation tower. For each of the examples below, we feed visual and audio features from two different videos but the correlation tower predicted the audio and the video features for each of the examples to be correlated as desired.}
\label{tab:corr_qual2}
\scriptsize
\begin{tabular}{l|c|c}
 FP         &    Example 1      &  Example 2  \\ \hline \hline
Video       &    \includegraphics[width=0.35\columnwidth]{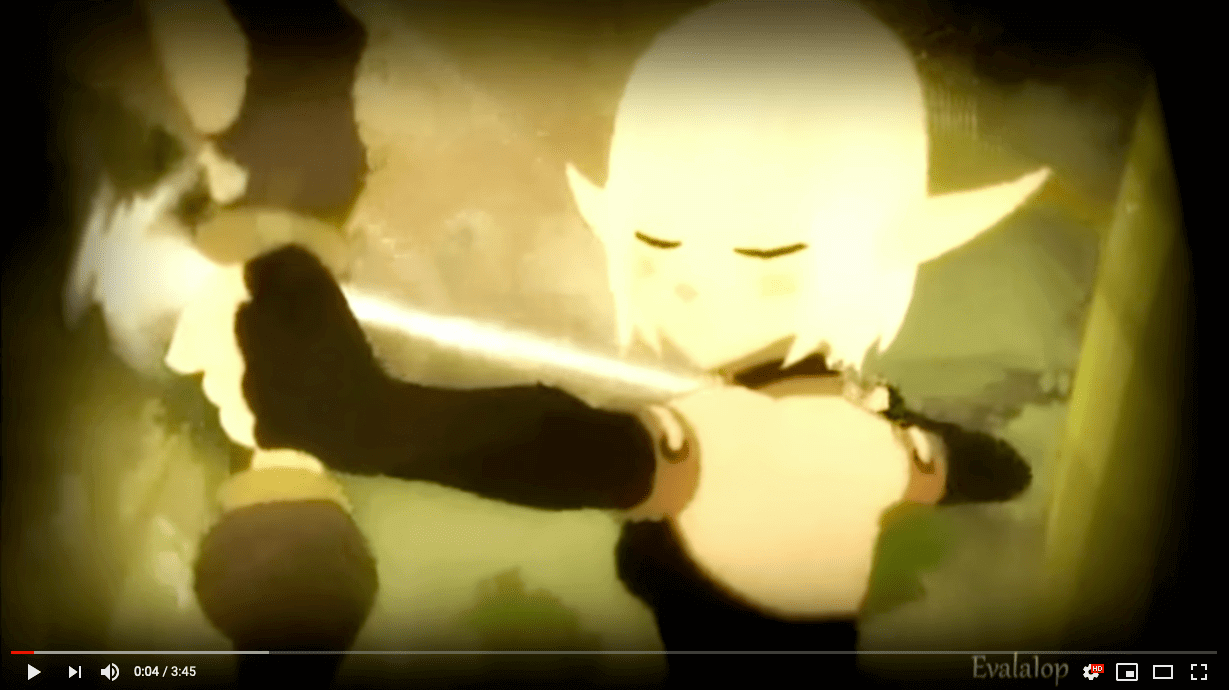}   &  \includegraphics[width=0.35\columnwidth]{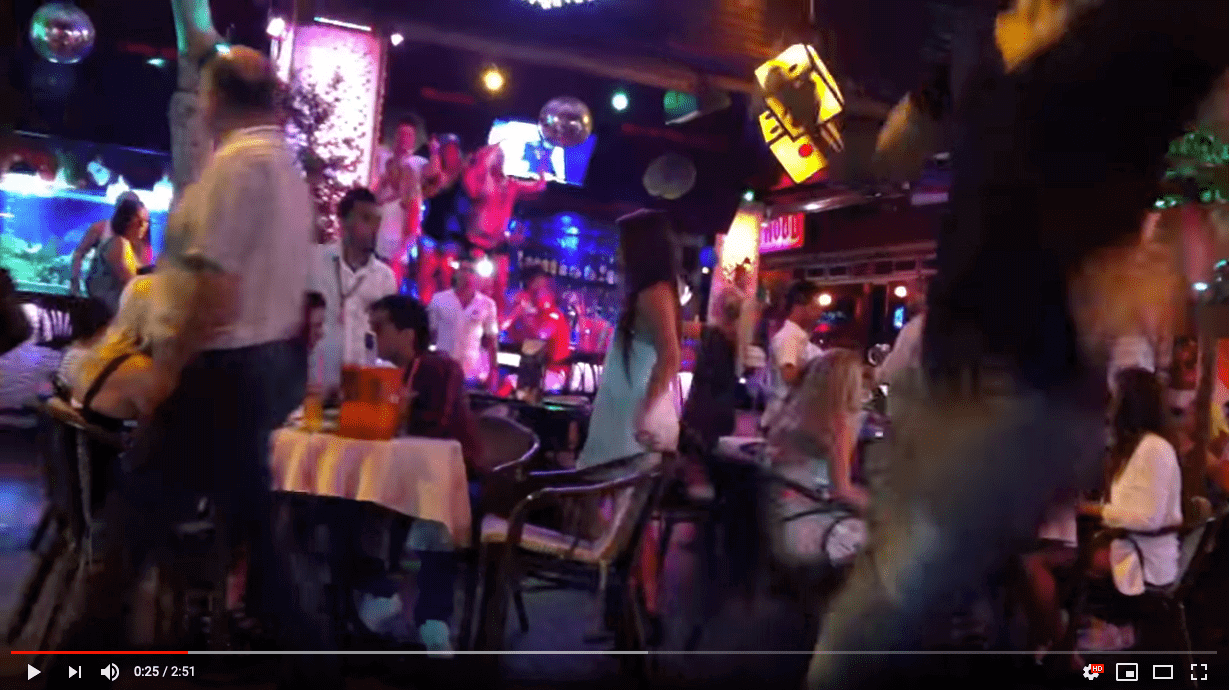}          \\ \hline\hline
Link(v) & bit.ly/32PptC8 & bit.ly/2TFaPtk \\ \hline \hline
Link(a) & bit.ly/2PLs0Iu & bit.ly/2TFjgoq \\ \hline \hline
Content(v) & Anime & Club Scene \\ \hline \hline
Content(a) & Cartoon Music & Music \\ \hline \hline
Prediction & Correlated & Correlated \\ \hline \hline
\end{tabular}
\vspace{-6mm}
\end{table}

We provide some examples from our trained correlation tower in Tables~\ref{tab:corr_qual1} and \ref{tab:corr_qual2}. In Table~\ref{tab:corr_qual1}, we have examples of visual and audio features from the same video but the model classifies them as uncorrelated. In example 1, we see that the video shows people skiing in the snow, whereas the audio is a background music unrelated to the video. Similarly, in example 2, video is an anime and the audio is an audio non-descriptive of the visual features. Thus, although the model is given an incorrect ground truth, it is able to identify the lack of correlation correctly. In Table~\ref{tab:corr_qual2}, we show examples where in each case, the visual and audio features were picked from two different videos but our correlation tower gave high correlation to them, as desired. In example 1, the visual content is anime, but the link for audio is a cartoon music. Example 2 is very interesting since in the video, we have a party scene with music --- the link for the audio contains a car show, but also contains party music. We see that in both these cases, the video and audio features, although from different links, were correlated and our correlation model was able to infer correctly.

\subsubsection{Error Analysis}
\label{sec:errAnalysis}
\begin{table}[hbtp]
\vspace{-8mm}
\centering
\caption{Error analysis of correlation tower. We show examples where the correlation tower makes mistakes.}
\label{tab:corr_err}
\scriptsize
\begin{tabular}{l|c|c}
 FN           &    Example 1      &  Example 2  \\ \hline \hline
Video       &    \includegraphics[width=0.35\columnwidth]{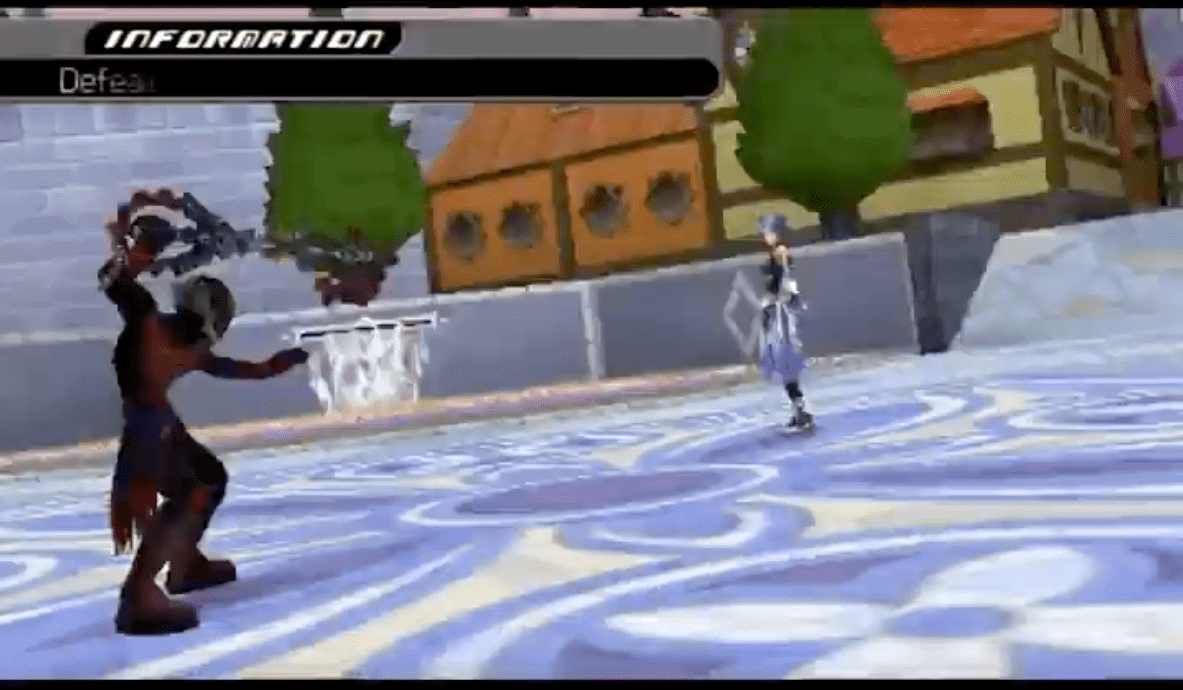}   &  \includegraphics[width=0.35\columnwidth]{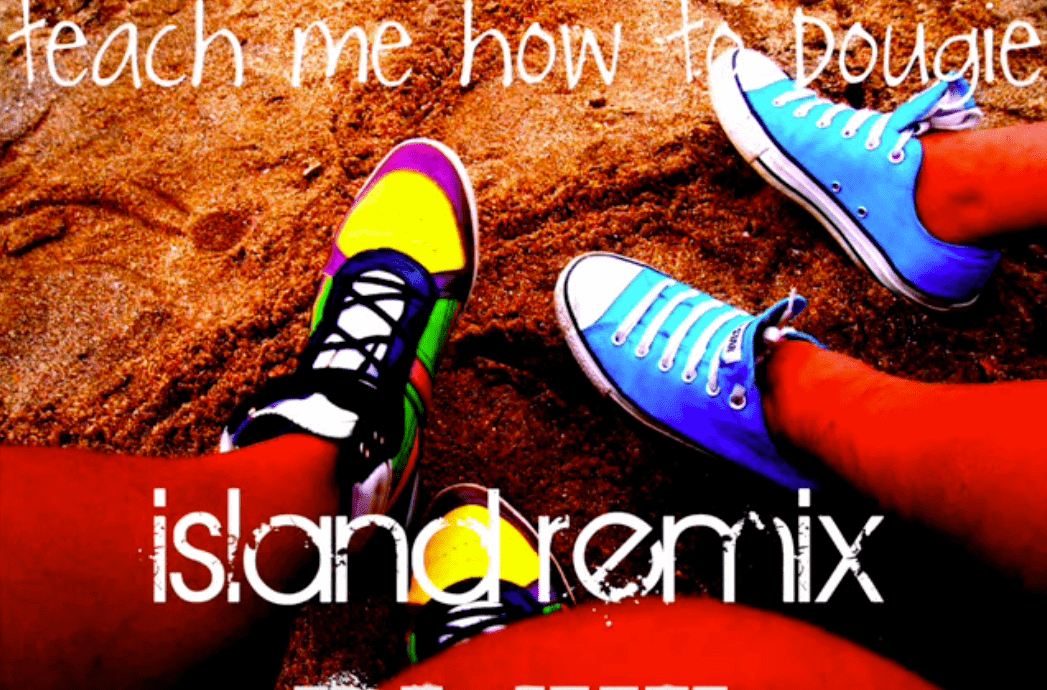}          \\ \hline\hline
Link & bit.ly/2uY4gK3 & bit.ly/2wsXSe5 \\ \hline \hline
Content(v) & Game & Poster \\ \hline \hline
Content(a) & Game Sound & Music \\ \hline \hline
Prediction & Uncorrelated & Correlated \\ \hline \hline
\end{tabular}
\vspace{-6mm}
\end{table}
We now look at examples where the correlation model's prediction match with the ground truth labels (see Table~\ref{tab:corr_err}). In example 1, the video is about a fighting game and the audio is game sound/music. The correlation tower predicts them to be uncorrelated which is incorrect. This may be because the model is confusing the sound with general music. In example 2, the video is a still poster with the name of the song whereas the audio is music. Although the video and audio seem uncorrelated, the correlation tower predicts them to be correlated. This may be attributed to the fact that many music videos contain such still pictures and the model learns that pattern.

Seeing the above examples, we believe that adding strong supervision to the model can improve the correlation tower's performance further.

\subsection{Video Categorization using Transformer}
\subsubsection{Quantitative Results}

\begin{table}[!t]
\centering
\caption{Error rate in GAP, Hit$@1$ and MAP scores in percentages ({\em note that lower is better}) at different hierarchical levels for different models for video categorization task on Youtube-8M using Transformer. E, L, G and C respectively denote early fusion, late fusion, gated and concatenation, while TM indicates the Transformer model. Improvement obtained by TM-CM-C model with respect to TM-L is mentioned in bold.}
\label{tab:youtube_model_perf}
\begin{tabular}{c||c|c|c|c}
  GAP       &   TM-E    &    TM-L         &   TM-CM-G     &   TM-CM-C\\ \hline \hline
Overall     &   9.18    &    8.67         &    8.65       &    8.49 \textbf{(0.18)}     \\ \hline
Level 0     &  5.11     &    4.81         &    4.8          &    4.72 \textbf{(0.09)}   \\ \hline
Level 1     &  7.00         &    6.59         &    6.61          &    6.52 \textbf{(0.07)}    \\ \hline
Level 2     &  9.10         &    8.58         &    8.63           &    8.50 \textbf{(0.08)}   \\ \hline
Level 3     &  12.23         &    11.42        &   11.34           &    11.27 \textbf{(0.15)}    \\ \hline
\end{tabular}
\begin{tabular}{c||c|c|c|c}
   MAP     &   TM-E       &    TM-L         &   TM-CM-G     &   TM-CM-C\\ \hline \hline
Overall     &  12.44           &    11.82        &   11.82            &  11.66 \textbf{(0.16)}    \\ \hline
Level 0     &  11.29           &    10.79        &   10.88             &  10.69 \textbf{(0.10)}   \\ \hline
Level 1     &   11.15          &    10.74        &   10.81            &   10.71 \textbf{(0.03)}   \\ \hline
Level 2     &  10.53           &    10.05        &    9.99            &   9.84 \textbf{(0.21)}     \\ \hline
Level 3     &  12.17           &    11.25        &   11.14            &   11.09 \textbf{(0.16)}     \\ \hline
\end{tabular}

\begin{tabular}{c||c|c|c|c}
   PERR         &   TM-E       &    TM-L         &   TM-CM-G     &   TM-CM-C \\ \hline \hline
Overall     &   11.93             &    11.43        &   11.32            &   11.25 \textbf{(0.18)}    \\ \hline
Level 0     &  7.16             &    6.85         &   6.66            &    6.68 \textbf{(0.17)}   \\ \hline
Level 1     &  8.73             &    8.45         &   8.27            &    8.24 \textbf{(0.21)}  \\ \hline
Level 2     &  12.88             &    12.31        &   12.24            &   12.17 \textbf{(0.14)}   \\ \hline
Level 3     &   16.30           &    15.36        &   15.19            &   14.87 \textbf{(0.49)}    \\ \hline
\end{tabular}

\begin{tabular}{c||c|c|c|c}
   Hit@1    &   TM-E    &    TM-L         &   TM-CM-G     &   TM-CM-C \\ \hline \hline
Overall     &  5.84         &    5.60         &   5.47            &   5.49 \textbf{(0.11)}   \\ \hline
Level 0     &  5.71         &    5.45         &   5.31              &   5.31  \textbf{(0.14)}   \\ \hline
Level 1     &  6.64         &    6.35         &   6.20              &   6.11 \textbf{(0.24)}    \\ \hline
Level 2     &  9.41         &    9.04         &   9.09              &   8.91 \textbf{(0.13)}    \\ \hline
Level 3     & 12.71          &    11.89        &   11.76            &   11.59 \textbf{(0.30)}   \\ \hline
\end{tabular}
\end{table}
\begin{table}[!t]
\centering
\caption{Error rate in GAP, Hit$@1$ and MAP scores in percentages ({\em note that lower is better}) for various top-level categories for video categorization task on Youtube-8M using Transformer. We show comparison for the best baseline model and our model. Gain captures the improvement of our model compared to the baseline.}
\label{tab:youtube_model_perf_cat}
\begin{tabular}{c||c|c|c|c|c|c|c|c}
  GAP       &   Electr.    &   Sports   &   Movie   &   Art     &   Transp.    &   Food    &   Games    &   Travel\\ \hline \hline
TM-L        &   13.72      &   2.19     &   3.31    &   2.50    &   7.16       &   13.81   &   3.98    &   12.92      \\ \hline
TM-CM-C     &   13.67      &   2.10     &   3.34    &   2.42    &   7.02       &   13.44   &   3.96    &   12.80      \\ \hline
Gain        &   0.05       &   0.09     &   -0.03   &   0.08    &   0.14       &   0.37    &   0.02    &   0.12      \\ \hline
\end{tabular}
\end{table}

We feed the outputs of the correlation tower into the cross modal model (explained in Section~\ref{sec:correlation}). Table~\ref{tab:youtube_model_perf} presents the evaluation results. For TM, we report results for 4 model variations: (i) E (Early Fusion), in which the visual and audio features are concatenated before feeding to the combined model, (ii) L (Late Fusion), in which we have separate towers for visual and audio features and combine them using a deep neural network at the end, (iii) CM-G (Cross Modal-Gated), where we supply the correlation tower output as a gate to the cross-modal layer, and (iv) CM-C (Cross Modal-Concatenated), where we concatenate the correlation tower output with the individual modality input in the cross-modal layer. All the results are an average of 5 test runs. For each evaluation metric, we also show level-based results. For this, we take $l^{th}$ level of the label hierarchy and evaluate on ground truth and predictions only for labels at that level. Examples for which there is neither prediction nor ground truth for a level are discarded for evaluation purposes.

Our first observation is that in all the experiments, late fusion achieves higher performance than early fusion. This indicates that understanding individual modalities before combining them gives improved performance. Our second observation is that introducing correlation gate (CM-G) to assist cross-modal dependency between individual modality towers further improves the performance. The model is further improved when we concatenate the correlation output to the cross-modal layer (CM-C). This may be possibly because when we concatenate the correlation output, we let the model learn the dependency between the correlation and the attention dependencies between the modalities. In other words, concatenation can be thought of as a generalization of the correlation gate, and is therefore giving better results in general. Another trend we observe is that the performance boost varies with the categories. Consider Table~\ref{tab:youtube_model_perf_cat} which shows performance in top-level categories. Categories such as ``Food'', ``Transport'', ``Travel'' and ``Sports'' gain the most from learning a correlation dependency between audio and video compared to categories like ``Games'' and ``Movie''. This follows intuition --- identifying relevant sounds of food, vehicles and cheering/sports related sound can be crucial in differentiating them from other categories. However, categories such as ``Movie'' has a variety of associated sounds, making it difficult for the model to identify them based on audio accurately.  Similarly, the performance boost increases going down the label hierarchy (see Table~\ref{tab:youtube_model_perf}). This is also intuitive as predicting a fine-grained label is harder and requires a better understanding of all available information.

\subsubsection{Qualitative Analysis}

Table~\ref{tab:qual_trans} shows the results of qualitative analysis of some representative examples for which transformer model with cross-modal and correlation tower information fares better than the baseline transformer model. For each Youtube video, along with the ground truth labels we show the top two predictions and their probabilities with the condition that the probability value should be at least 0.3. While baseline model fails to predict the classes correctly, our model with cross-modal correlation information is able to predict them correctly. Of particular interest is example 3 where our model is able to predict the classes correctly even in low-light conditions.

\begin{table*}[hbtp]
\centering
\caption{Qualitative analysis for some example videos in Youtube 8M comparing the performance of Transformer-CrossModal with baseline Transformer model. Abbreviations VG, Mov, Adv, Int, Anim, Trans, Spo and BB in table stand for VideoGame, MovieOrTV, Adventure, Interactive, Animation, Transport, Sports and BallBat respectively.}
\label{tab:qual_trans}
\scriptsize
\begin{tabular}{l|c|c|c|c}
            &    Example 1      &  Example 2      &   Example 3     &   Example 4 \\ \hline \hline
Video       &    \includegraphics[width=0.2\columnwidth]{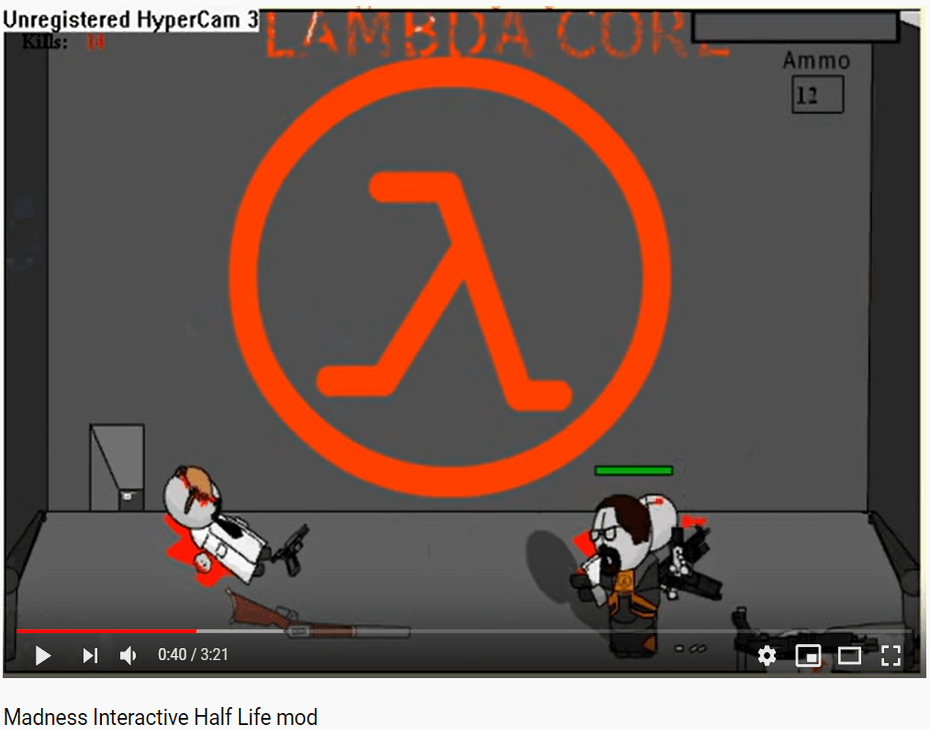}   &  \includegraphics[width=0.2\columnwidth]{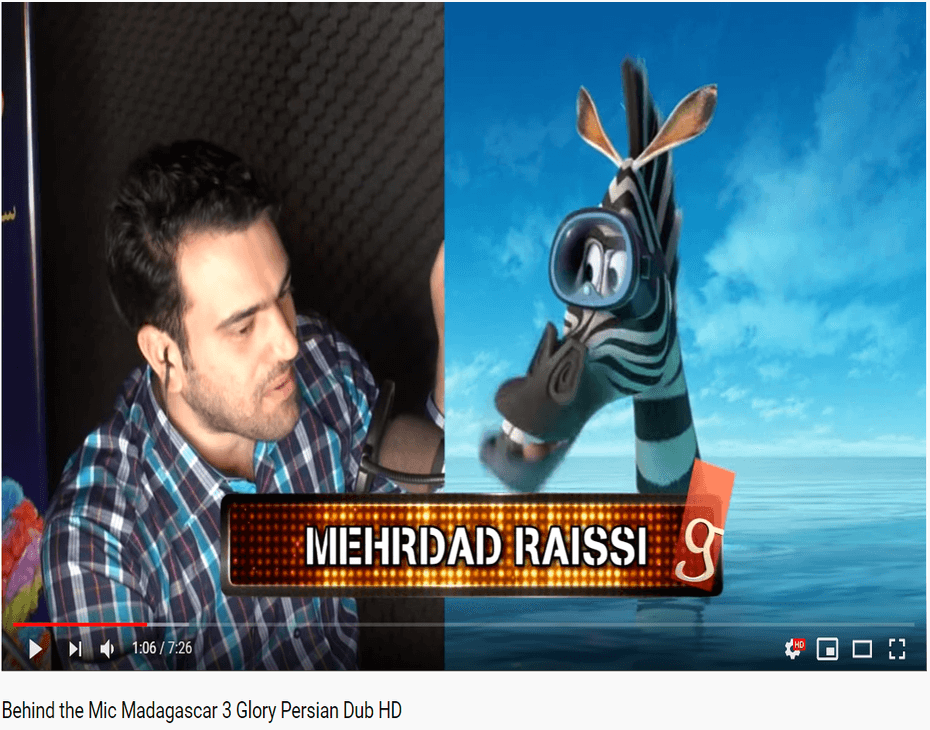}   &   \includegraphics[width=0.2\columnwidth]{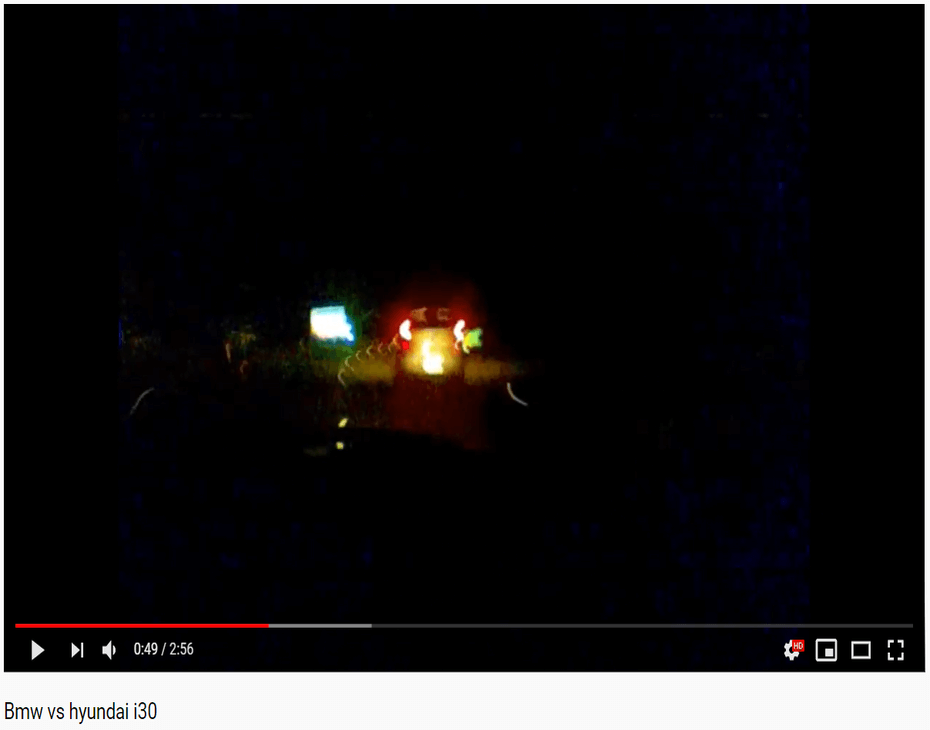}       &   \includegraphics[width=0.2\columnwidth]{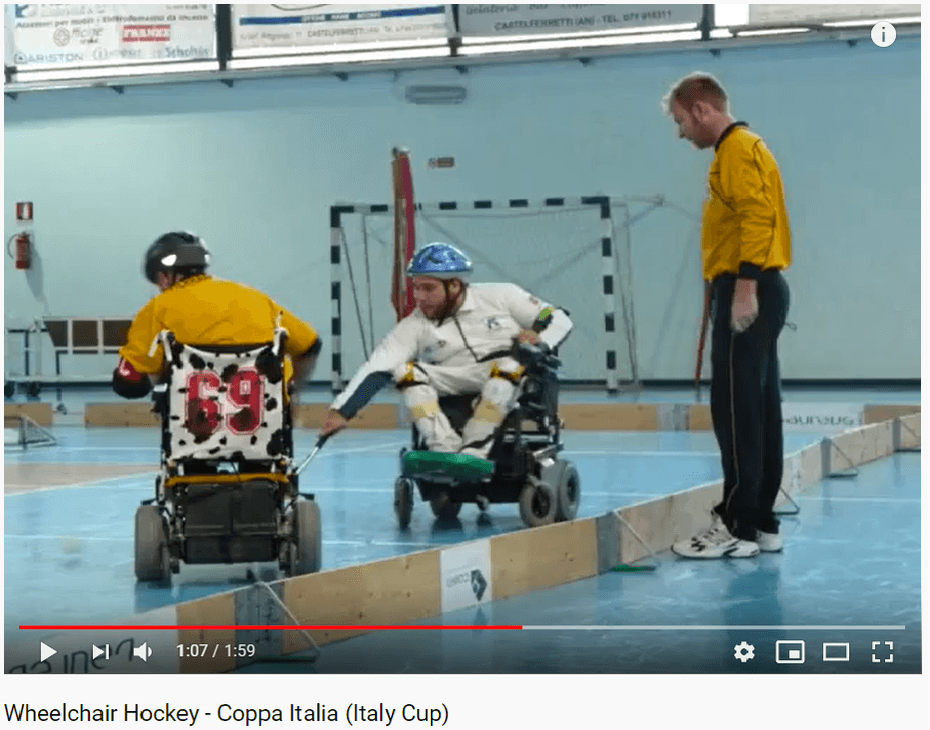}           \\ \hline\hline
Links & bit.ly/38rSzJc & bit.ly/2IzUMId & bit.ly/2VQ1TUn & bit.ly/2PQDryJ \\ \hline \hline
      \multirow{4}{*}{GT} 
      		& Game & Mov & Trans &Spo\\ 
      		& Game:VG & Mov:Adv & Trans:Land  &Spo:Ball \\ 
      		& Game:VG:Adv & Mov:Adv:Anim & Trans:Land:Car &Spo:Ball:BB\\
      		& Game:VG:Adv:Int & & &Spo:Ball:BB:Hoc\\ \cline{1-5}\hline\hline
      \multirow{2}{*}{TM-L} 
      		& Mov: 0.94 & Game:VG: 0.62  & Art: 0.33 & Trans:Land: 0.78\\
      		& Mov:Adv:Anim: 0.93& Game: 0.56 &  & Trans:Land:Car: 0.77\\  \cline{1-5}\hline\hline
      \multirow{2}{*}{TM-CM-C} 
      		& Game:VG: 0.97 & Mov:Adv:Anim: 0.76 & Trans: 0.42 & Spo: 0.55\\ 
      		& Game:VG:Adv:Int: 0.9 & Mov: 0.7 & Trans:Land:Car: 0.38 & Trans:Land: 0.48\\  \cline{1-5}
 \hline
\end{tabular}
\vspace{-4mm}
\end{table*}

\subsubsection{Error Analysis}
We now look at examples in Table~\ref{tab:err_trans} where the model's top prediction didn't match with the ground truth labels:
\begin{itemize}
\item In example 1 a person is comparing different smartphone cameras while clicking pictures and videos in a city. The assigned ground truth labels are ``Electronics'', ``Gadgets'' and ``Phones'' even though phone is not in the video. Since it's a photography related video, we note that both the baseline and TM-CM-C models predict the it as an ``Art'' with the latter predicting it with a higher probability. Interestingly, TM-CM-C model also classifies it as ``Transport'' because of the vehicles appearing in the video.

\item In example 2 we observe that the topmost prediction for both the models are ``Game'' while the actual video is about food items. However, we observe that for TM-CM-C model the prediction probability of it being a ``Game'' is closely followed by the class ``Food'' showing it is actually able to recognize the true class to some extent. 

\item Example 3 is a video of a guy reviewing a video game. However, throughout the entire duration of the video the video game is never played and the video remains pretty much the same from frame to frame. Both the models predict the class as ``Electronics'' and ``Gadget'', which is probably because there is a smart phone lying on the table in the video. We also note that throughout the video the person just talks about the video game --- so, we hope that having subtitles or text metadata would help us improve the prediction.

\item Finally, for example 4 the top predictions for both the models TM-L and TM-CM-C is ``Video Game", with the baseline model predicting it with a higher probability. However the actual video is a tutorial on how to use the software `Sprite' to make electronic art. Moreover, when we watch the Youtube video, the graphics resemble that of a video game leading the models to confuse it with a video game. The transformer model with cross-modal and correlation gate still classifies it as an `Art' video albeit with a low probability which shows it can recognize the correct class to some extent. Similar to example 3, having text metadata/subtitles would help in this case, since the person is talking about steps to create art in the video.
\end{itemize}
\begin{table*}[hbtp]
\centering
\caption{Error analysis for some example videos in Youtube 8M comparing the performance of Transformer-CrossModal with baseline Transformer model. Abbreviations Electr, FI, VG, Adv and Act in table stand for Electronics, FoodItems, VideoGame, Adventure and Action respectively.}
\label{tab:err_trans}
\scriptsize
\begin{tabular}{l|c|c|c|c}
            &    Example 1      &  Example 2      &   Example 3     &   Example 4 \\ \hline \hline
Video       &    \includegraphics[width=0.22\columnwidth]{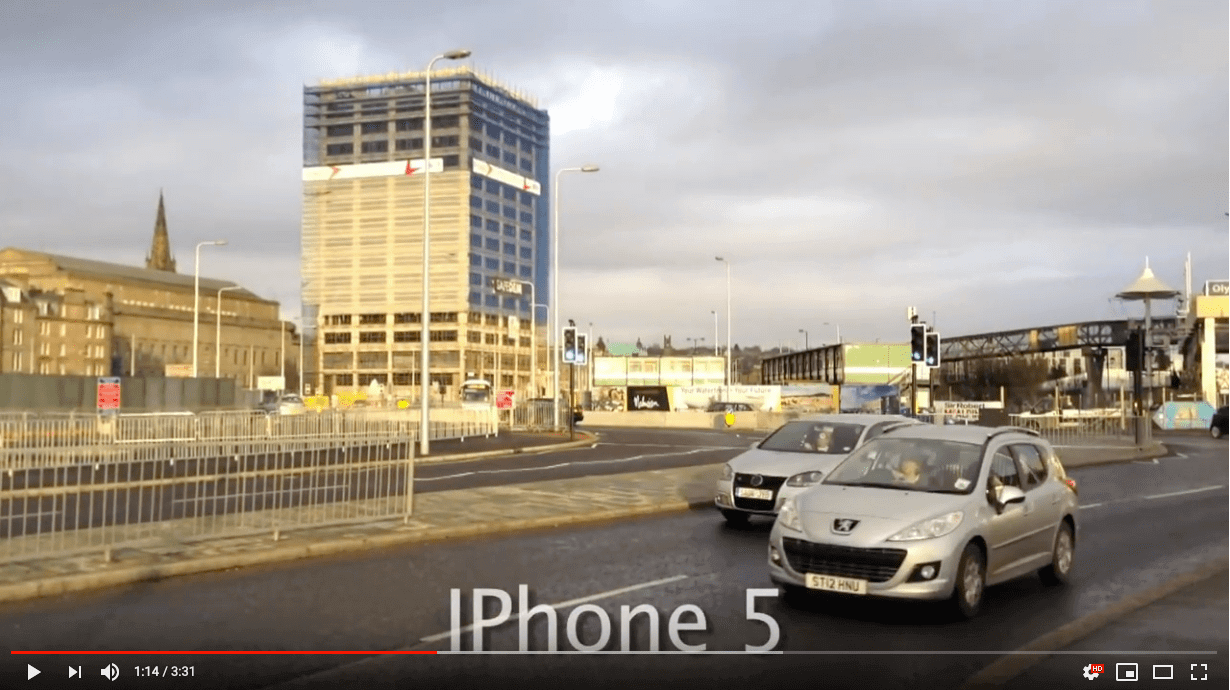}   &  \includegraphics[width=0.22\columnwidth]{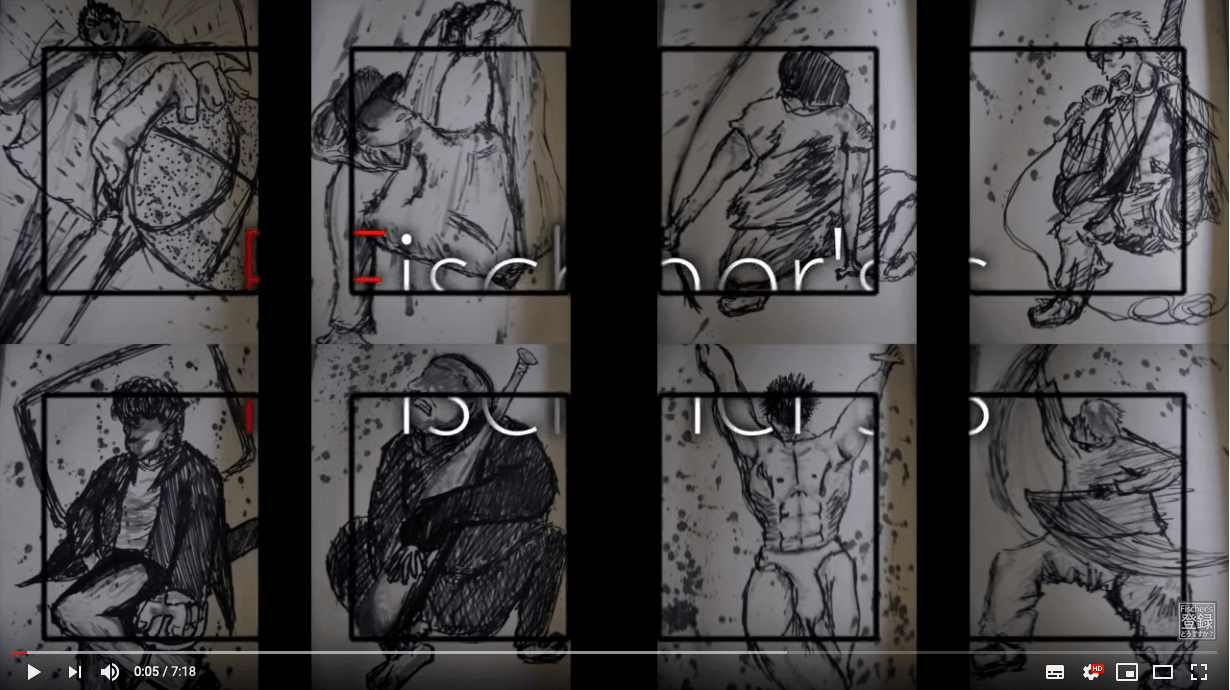}   &   \includegraphics[width=0.2\columnwidth]{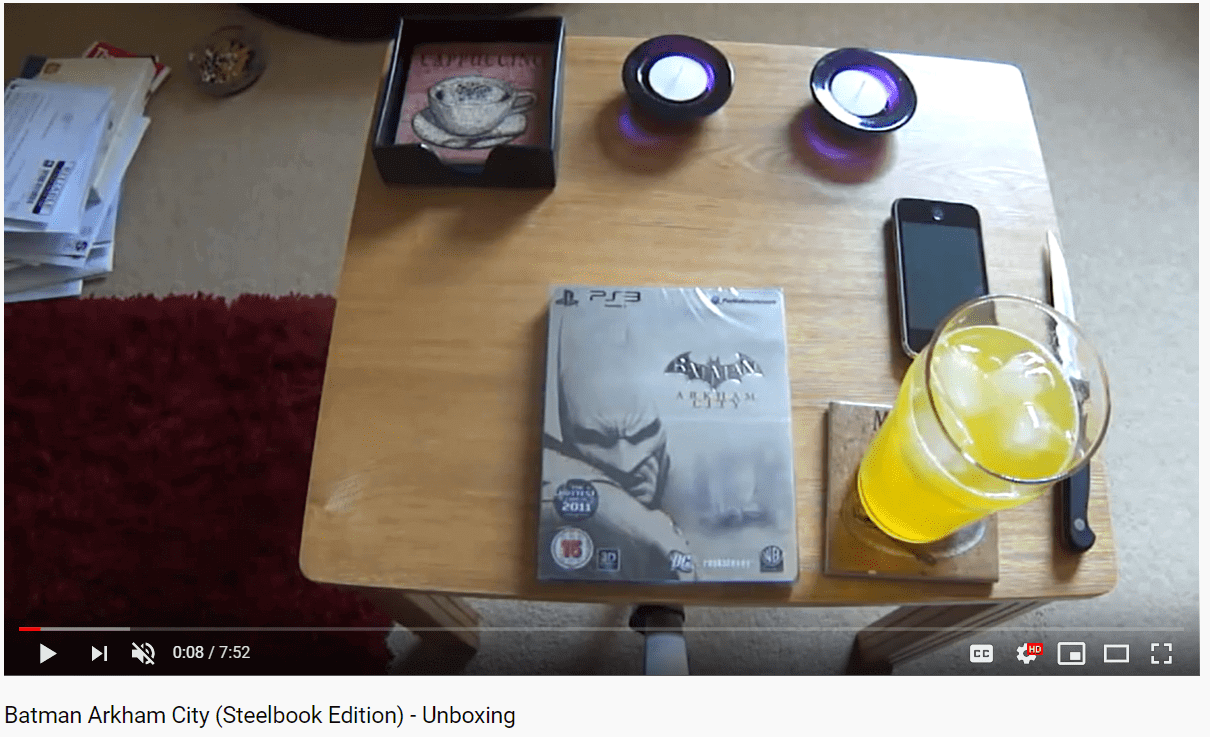}       &   \includegraphics[width=0.16\columnwidth]{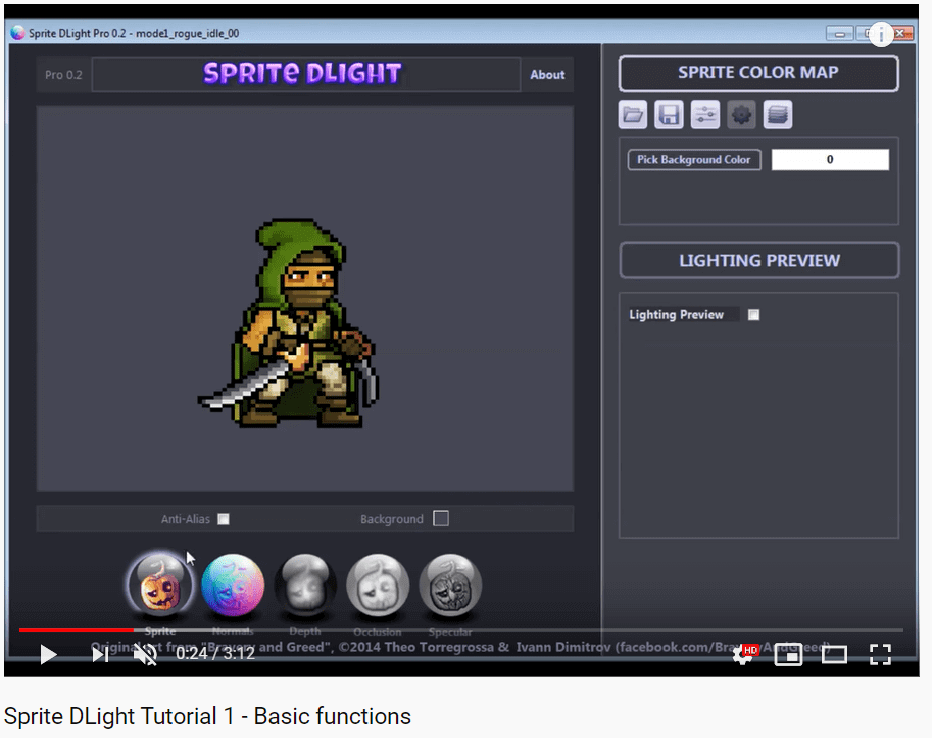}           \\ \hline\hline
Links & bit.ly/3cxDEAo & bit.ly/2xevFrU & bit.ly/32QVOsy & bit.ly/2TundNX \\ \hline \hline
      \multirow{4}{*}{GT} 
      		& Electr. & Food & Game &Art\\ 
      		& Electr:Gadget & Food:FI & Game:VG  & \\ 
      		& Electr:Gadget:Phone & Food:FI:Dessert & Game:VG:Adv &\\
      		&  & Food:Recipe & Game:VG:Adv:Act &\\ \cline{1-5}\hline\hline
      \multirow{2}{*}{TM-L} 
      		& Art: 0.47 & Game: 0.74  & Electr.: 0.96 &  Game:VG: 0.74\\
      		& Art:Photo: 0.39& Game:VG: 0.14 &  Electr.:Gadget: 0.97 & Electr.: 0.1\\  \cline{1-5}\hline\hline
      \multirow{2}{*}{TM-CM-C} 
      		& Art: 0.97 & Game: 0.24 & Electr.: 0.77 & Game:VG: 0.34\\ 
      		& Transport: 0.4 & Food: 0.23 & Electr.:Gadget: 0.84 & Art: 0.13\\  \cline{1-5}
 \hline
\end{tabular}
\vspace{-4mm}
\end{table*}

\subsection{Video Categorization using NetVLAD}
\subsubsection{Quantitative Results}
Table~\ref{tab:youtube_model_perf_netv} presents the evaluation results of NetVLAD model with and without our novel cross-modal layer and correlation tower. For our NetVLAD (NV) model we had a cluster size of 128 for visual modality and 64 clusters for audio modality. For NV, we report results for 4 model variations: (i) E (Early Fusion), in which the visual and audio features are concatenated before feeding to the combined model, (ii) L (Late Fusion), in which we have separate towers for visual and audio features and combine them using a deep neural network at the end, (iii) CM-G (Cross Modal-Gated), where we supply the correlation tower output as a gate to the cross-modal layer, and (iv) CM-C (Cross Modal-Concatenated), where we concatenate the correlation tower output with the individual modality input in the cross-modal layer. All the results are an average of 5 test runs. For each evaluation metric, we also show level-based results. For this, we take $l^{th}$ level of the label hierarchy and evaluate on ground truth and predictions only for labels at that level. Examples for which there is neither prediction nor ground truth for a level are discarded for evaluation purposes.

We first note that the gain we get from our cross-modal layer is significantly higher than our gain in Transformer. There can be two reasons for this. Firstly, the headroom for improvement in the NetVLAD model could be higher than that in Transformer model, since the latter has a stronger baseline model. Secondly, NetVLAD model is a clustering-based algorithm --- in general clustering-based methods gain more from constrained optimization~\cite{zhang2019framework}, so the constraint from one modality could be helping the prediction in the other modality.

Our second observation is that in all the experiments, late fusion achieves higher performance than early fusion. This is consistent with our results with Transformers (main paper) and indicates that understanding individual modalities before combining them gives improved performance. 

Our third observation, again consistent with our Transformers results, is that introducing correlation gate (CM-G) to assist cross-modal dependency between individual modality towers further improves the performance. The model is further improved when we concatenate the correlation output to the cross-modal layer (CM-C). This may be possibly because when we concatenate the correlation output, we let the model learn the dependency between the correlation and the attention dependencies between the modalities.

We now consider Table~\ref{tab:youtube_model_perf_netv_cat} which shows performance in top-level categories. In Transformers (main paper), we had observed that categories such as ``Food'', ``Transport'', ``Travel'' and ``Sports'' gained the most from learning a correlation dependency between audio and video compared to categories like ``Games'' and ``Movie''. In NetVLAD, we see that the trend is consistent for ``Travel'', ``Transport'', ``Food'' and ``Sports''. However, we observe that here we also get significant gain for categories such as ``Games'' and ``Movie''. This shows that the NetVLAD model is gaining for more categories by using the cross-modal layer.
\begin{table}[!t]
\centering
\caption{Error rate in GAP, Hit$@1$ and MAP scores in percentages ({\em note that lower is better}) at different hierarchical levels for different models for video categorization task on Youtube-8M for NetVLAD model. E, L, G and C respectively denote early fusion, late fusion, gated and concatenation, while NV indicates the NetVLAD model. Improvement obtained by NV-CM-C model with respect to NV-L is mentioned in bold.}
\label{tab:youtube_model_perf_netv}
\begin{tabular}{c||c|c|c|c}
  GAP       &   NV-E     &    NV-L         &   NV-CM-G     &   NV-CM-C\\ \hline \hline
Overall     &   10.81    &    10.61         &   9.90    &    9.75 \textbf{(0.86)}     \\ \hline
Level 0     &   6.53     &    6.33         &   5.65     &    5.61 \textbf{(0.72)}   \\ \hline
Level 1     &   8.13     &    8.03         &   7.38       &    7.26 \textbf{(0.77)}    \\ \hline
Level 2     &   10.41    &    10.25         &  9.57       &    9.43 \textbf{(0.82)}   \\ \hline
Level 3     &   14.34    &    13.78        &   13.12   &    12.85 \textbf{(0.93)}    \\ \hline
\end{tabular}
\begin{tabular}{c||c|c|c|c}
   MAP      &   NV-E      &    NV-L         &   NV-CM-G     &   NV-CM-C\\ \hline \hline
Overall     &   13.46     &    13.14        &    12.58    &  12.43 \textbf{(0.71)}    \\ \hline
Level 0     &   13.52     &    13.51        &    12.18     &  12.32 \textbf{(1.19)}   \\ \hline
Level 1     &   12.36     &    12.19        &    11.36   &   11.32 \textbf{(0.87)}   \\ \hline
Level 2     &   11.14     &    10.94        &   10.48   &   10.31 \textbf{(0.63)}     \\ \hline
Level 3     &   13.73     &    13.07        &    12.63  &   12.48 \textbf{(0.59)}     \\ \hline
\end{tabular}

\begin{tabular}{c||c|c|c|c}
   PERR     &   NV-E        &    NV-L         &   NV-CM-G     &   NV-CM-C \\ \hline \hline
Overall     &   13.99       &     13.85       &   13.17  &   12.99 \textbf{(0.86)}    \\ \hline
Level 0     &    8.98       &       8.78      &  8.22     &    8.14 \textbf{(0.64)}   \\ \hline
Level 1     &   10.30       &      10.29       &  9.69     &   9.50  \textbf{(0.79)}  \\ \hline
Level 2     &   14.37       &     14.38       &  13.83    &  13.71  \textbf{(0.67)}   \\ \hline
Level 3     &   17.88       &      17.32      & 17.19    &   16.88 \textbf{(0.44)}    \\ \hline
\end{tabular}

\begin{tabular}{c||c|c|c|c}
   Hit@1    &   NV-E    &    NV-L         &   NV-CM-G     &   NV-CM-C \\ \hline \hline
Overall     &   7.14    &     6.95        &  6.61             &    6.56\textbf{(0.39)}   \\ \hline
Level 0     &   7.17    &    6.97         &  6.57               &   6.48  \textbf{(0.49)}   \\ \hline
Level 1     &   7.87    &    7.79         & 7.34                &    7.13\textbf{(0.66)}    \\ \hline
Level 2     &   10.58   &    10.56         &  10.20               &    10.11\textbf{(0.45)}    \\ \hline
Level 3     &   14.24   &   13.13         &   13.15            &    12.91\textbf{(0.22)}   \\ \hline
\end{tabular}
\end{table}
\begin{table}[!t]
\centering
\caption{Error rate in GAP, Hit$@1$ and MAP scores in percentages ({\em note that lower is better}) for various top-level categories for video categorization task on Youtube-8M. We show comparison for the best baseline model and our model. Gain captures the improvement of our model compared to the baseline.}
\label{tab:youtube_model_perf_netv_cat}
\begin{tabular}{c||c|c|c|c|c|c|c|c}
  GAP       &   Electr.    &   Sports   &   Movie   &   Art     &   Transp.    &   Food    &   Games    &   Travel\\ \hline \hline
NV-L        &   15.42      &   3.15     &   4.94    &   2.88    &   8.43       &   15.12   &   5.35    &   16.60      \\ \hline
NV-CM-C     &   15.49      &   2.85     &   4.17    &   2.64    &   7.90       &   14.53   &   4.76    &   14.30      \\ \hline
Gain        &   -0.07      &   0.30     &   0.77    &   0.24    &   0.53       &   0.59    &   0.61    &   2.30      \\ \hline
\end{tabular}
\end{table}

\subsubsection{Qualitative Analysis}

\begin{table*}[hbtp]
\centering
\caption{Qualitative analysis for some example videos in Youtube 8M comparing the performance of NetVLAD-CrossModal with baseline NetVLAD model. Abbreviations VG, Adv, Anim and Trans in table stand for VideoGame, Adventure, Animation and Transport respectively.}
\label{tab:qual}
\scriptsize
\begin{tabular}{l|c|c|c|c}
            &    Example 1      &  Example 2      &   Example 3     &   Example 4 \\ \hline \hline
Video       &    \includegraphics[width=0.2\columnwidth]{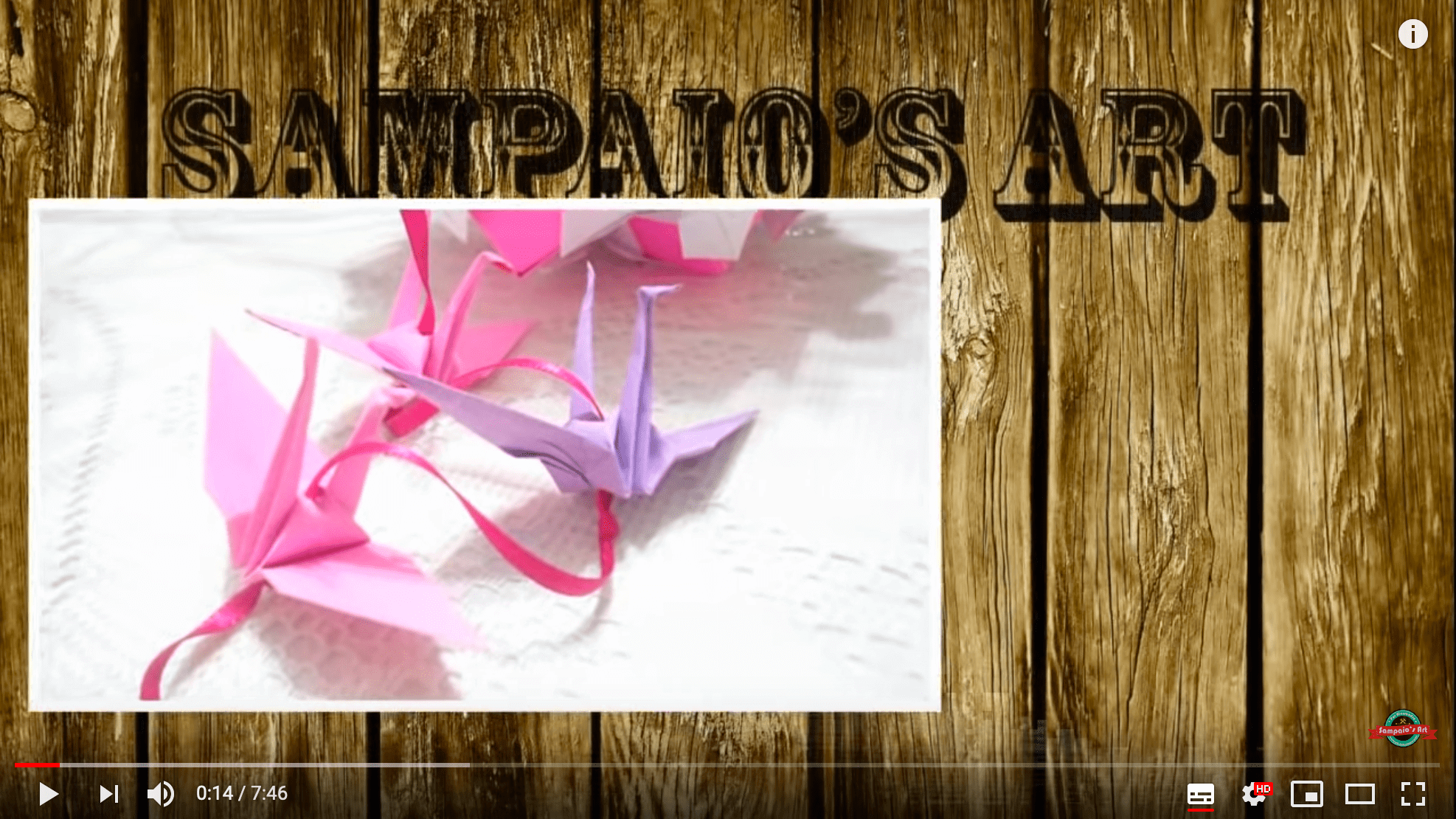}   &  \includegraphics[width=0.2\columnwidth]{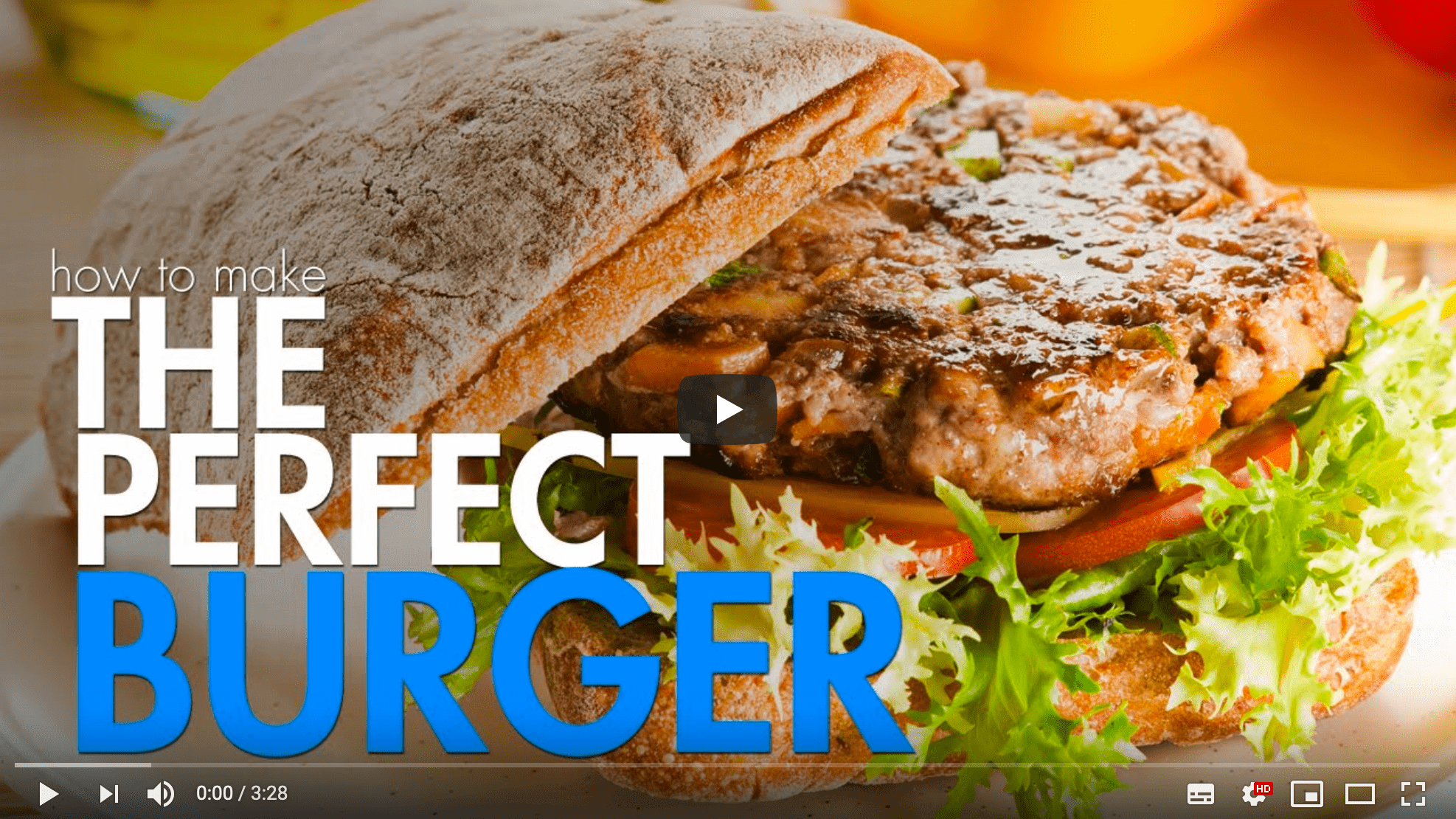}   &   \includegraphics[width=0.2\columnwidth]{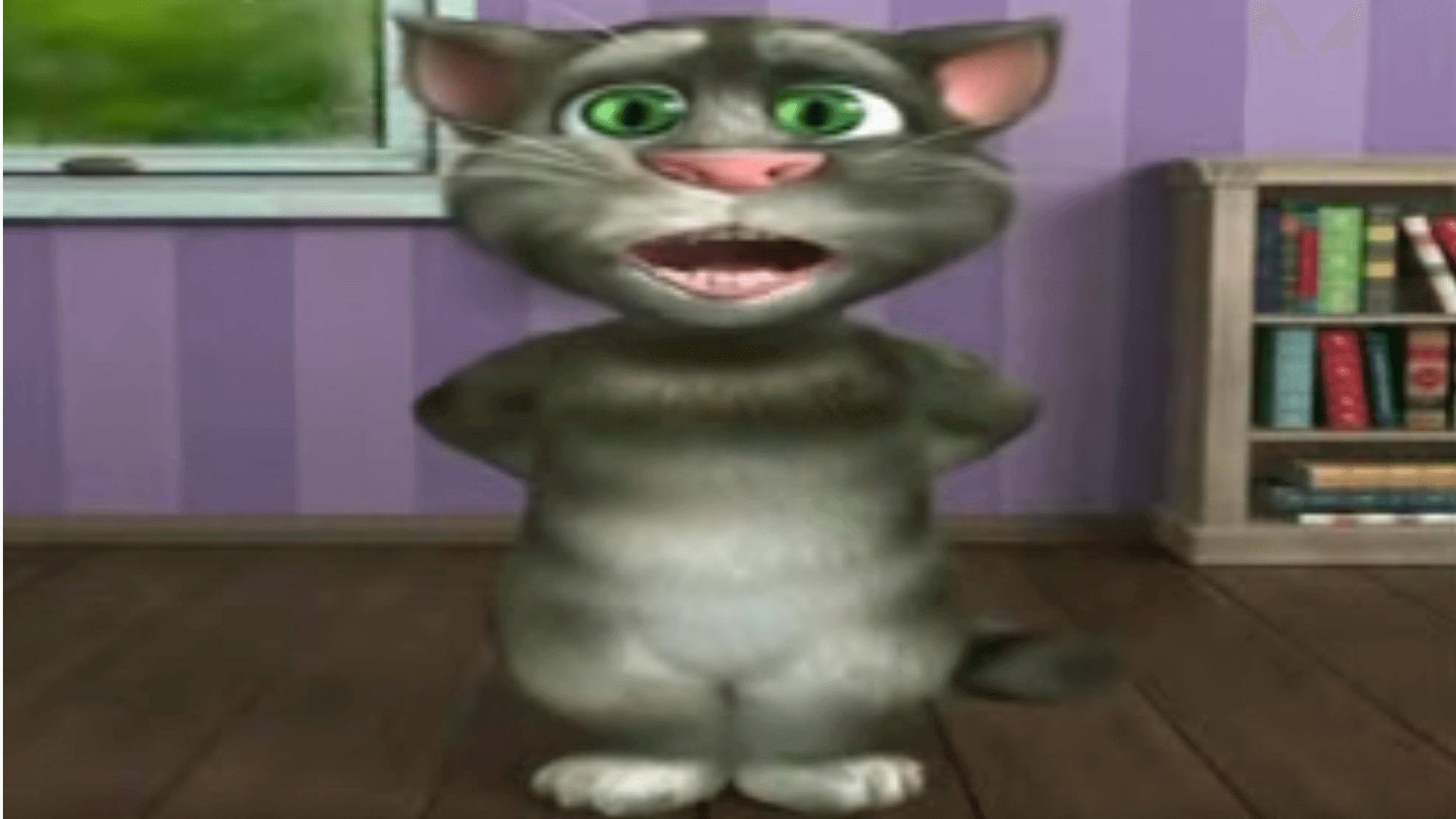}       &   \includegraphics[width=0.2\columnwidth]{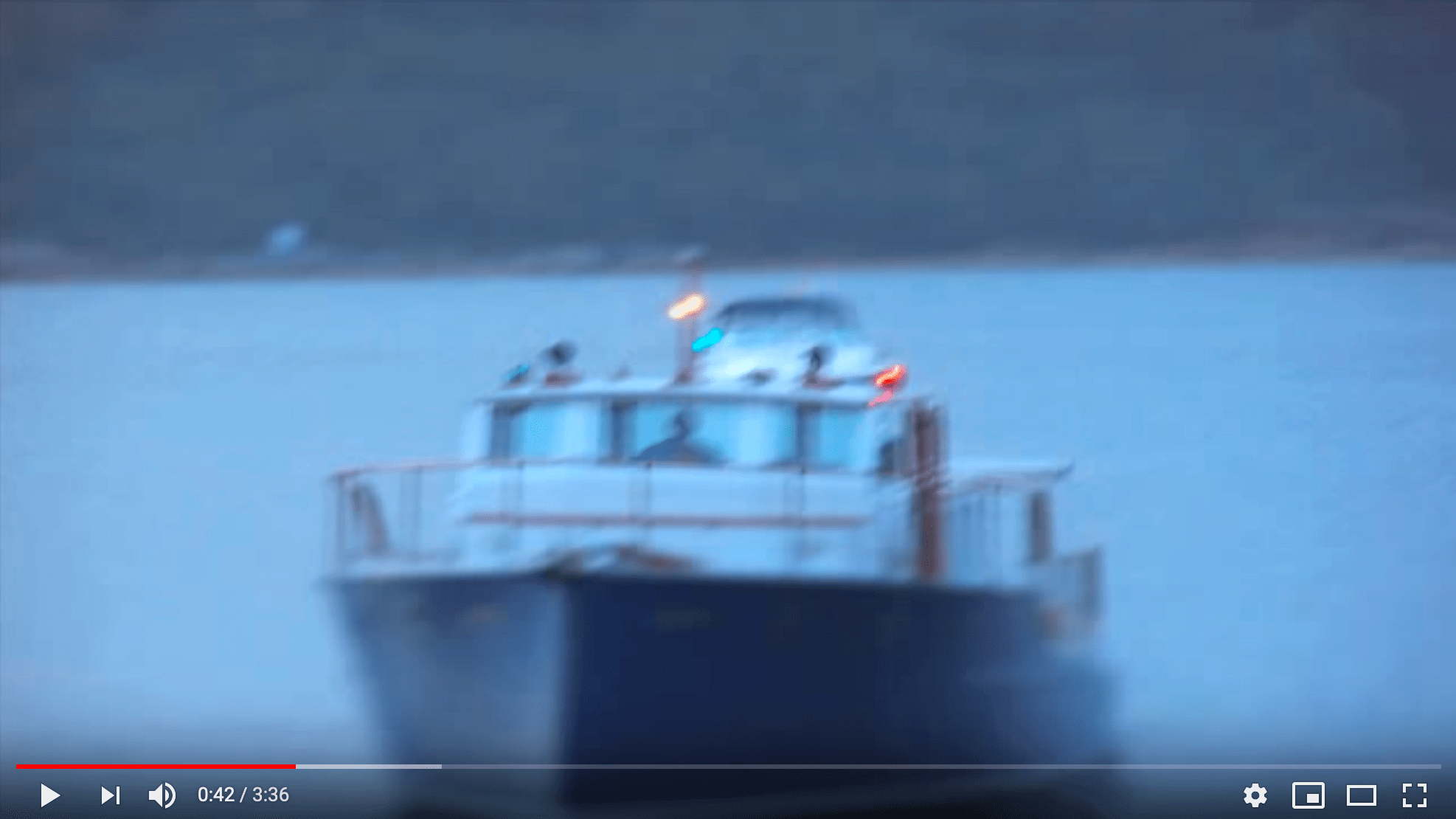}           \\ \hline\hline
Links & bit.ly/2Q2cx6R & bit.ly/2TE9Fzn & bit.ly/2VY5HTN & bit.ly/2xpkoVK \\ \hline \hline
      \multirow{4}{*}{GT} 
      		& Art & Food & MovieOrTV &Transport\\ 
      		& Art:Painting & Food:Recipe & MovieOrTV:Adv  &Trans:Water \\ 
      		&  & Food:FI & MovieOrTV:Adv:Anim &Trans:Water:Ship\\
      		&  & Food:FI:Sandwich & &\\ \cline{1-5}\hline\hline
      \multirow{2}{*}{NV-L} 
      		& Game: 0.7 & Art: 0.23  & Animal: 0.7 & Music:Concert: 0.66\\
      		& Game:VG: 0.57& Game: 0.22 &  Animal:Land: 0.6 & Music:Concert:DJ: 0.51\\  \cline{1-5}\hline\hline
      \multirow{2}{*}{NV-CM-C} 
      		& Art: 0.14 & Food: 0.41 & MovieOrTV:Adv: 0.69 & Trans:Water:Ship: 0.27\\ 
      		& Art:Painting: 0.12 & Elec:Gadget: 0.24 & MovieOrTV:Adv:Anim: 0.71 & Trans:Water: 0.26\\  \cline{1-5}
 \hline
\end{tabular}
\vspace{-4mm}
\end{table*}

Table~\ref{tab:qual} shows the results of qualitative analysis of some representative examples for which NetVLAD model with cross-modal and correlation tower information fares better than the baseline NetVLAD model. For each Youtube video, along with the ground truth labels we show the top two predictions and their probabilities:
\begin{itemize}
    \item In example 1, a person is teaching an art class. This example is particularly tricky as the background music at the beginning resembles that of a video game. Further, the art form is not traditional painting and uses a water glass. The baseline NetVLAD predicts video game, likely due to the background sound. The cross-modal model is able to make the top 2 predictions correctly, identifying that it is an art video.
    \item Example 2 is an instructional video from a chef on burger recipe. It is a difficult video for categorization as the burger doesn't appear throughout the video and the background music can be confusing. The baseline model predicts art and game for this video whereas our cross modal model predicts food as the most likely category.
    \item In example 3, there is an animated cat which imitates a person's singing. Note that we only tag videos of real animals with the label ``Animal''. The baseline model isn't able to differentiate between real and animated cat and predicts animal as the tag for this video. Our model, however, makes accurate prediction that the video is animated.
    \item Example 4 contains a still image of a ship with background music. The baseline model is affected by the background music and predicts the video to be a music video. Our model, however, filters out the music and focuses on the ship, thus predicting the ground truth labels correctly. There are several such examples where baseline model falsely predicts music because of the sound.
\end{itemize}

\subsubsection{Error Analysis}
We now look at examples in Table~\ref{tab:err_netvlad} where the model's top prediction didn't match with the ground truth labels:
\begin{itemize}
\item In example 1, we see a video of a person entering a Lamborghini and going around a race track to test its speed. Given the Lamborghini, the ground truth tags this video with Italian car. However, baseline NetVLAD and our model also classifies it as ``Racing'', likely because of the racing track and the surroundings. 

\item Example 2 is a difficult example containing a wide range of images. It's showing a food factory and various buildings and infrastructure related to it. Thus, the ground truth is ``Food''. Just looking at the video and hearing the sound which contains music, it is very difficult to make out that the factory manufactures food. Upon verification, we see two scenes at times 2:33 and 3:03 in the video showing ice cream. However, there are several scenes (including 1:13 and 2:46) where we see cars. The baseline and our model classifies the video as ``Transport'' and ``Transport:Land:Car'' because of these scenes.

\item Example 3 is a case of faulty ground truth. The video is a news clip showing the arrest of an outlaw. Although the ground truth is ``Animal:Bird:Fowl'', our models correctly categorize the video as ``MovieOrTV:News''

\item Finally, example 4 shows a video with scenes taken from a TV show with background music. The ground truth is ``MovieOrTV''. However, the baseline model predicts ``Music'' and ``Transport''. Our cross-modal model has better predictions with ``Music'' and ``MovieOrTV'' although both the models get the top predictions wrong. In several examples (some shown in Qualitiative Analysis section), the cross-modal model is able to filter out the background music to classify correctly but in some difficult examples such as this which resemble more to music videos, the model still falls short of identifying the video category correctly.
\end{itemize}
\begin{table*}[hbtp]
\centering
\caption{Error analysis for some example videos in Youtube 8M comparing the performance of NetVLAD-CrossModal with baseline NetVLAD model. Abbreviations Trans and FI in table stand for Transport and FoodItems respectively.}
\label{tab:err_netvlad}
\scriptsize
\begin{tabular}{l|c|c|c|c}
            &    Example 1      &  Example 2      &   Example 3     &   Example 4 \\ \hline \hline
Video       &    \includegraphics[width=0.22\columnwidth]{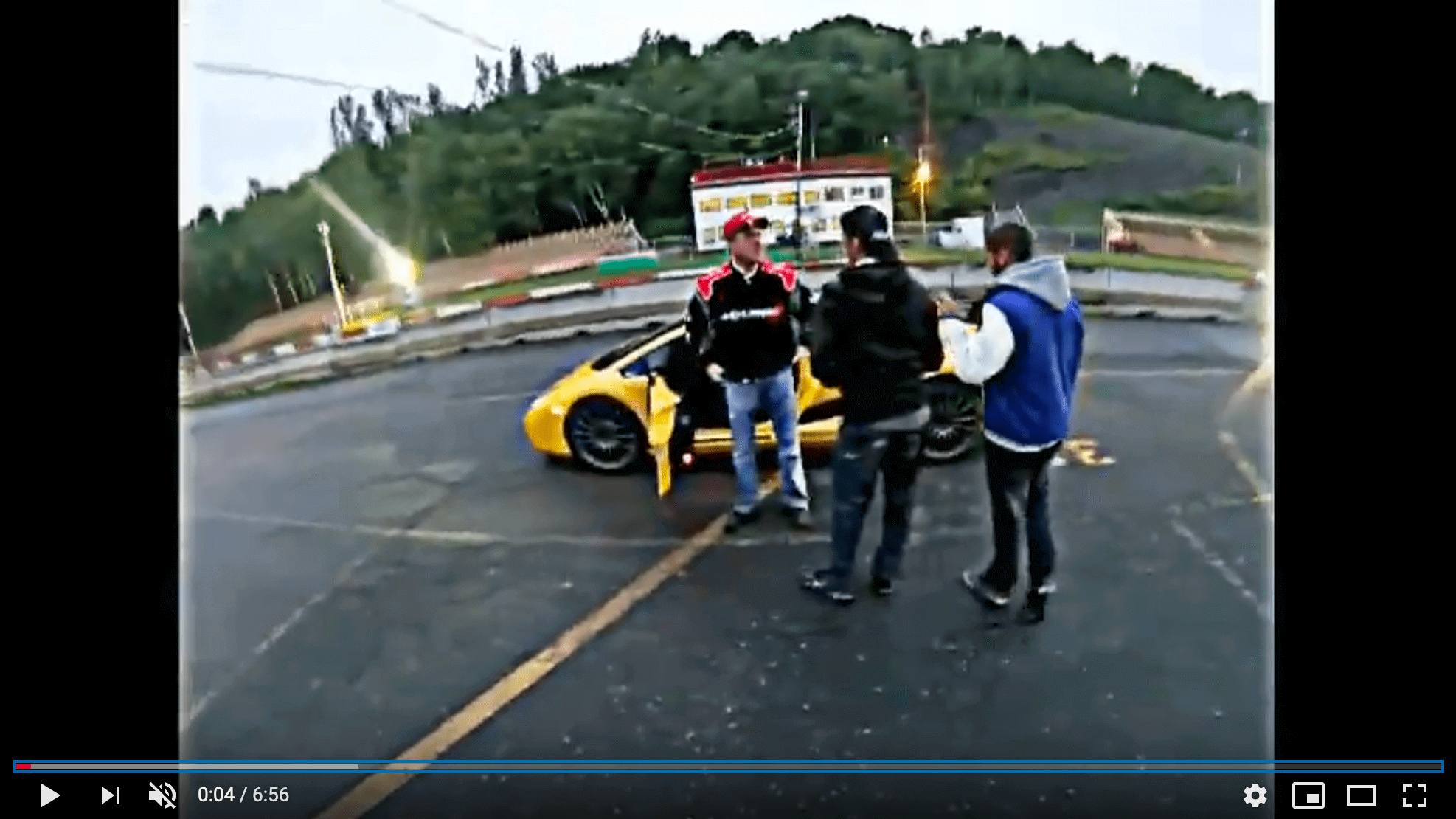}   &  \includegraphics[width=0.22\columnwidth]{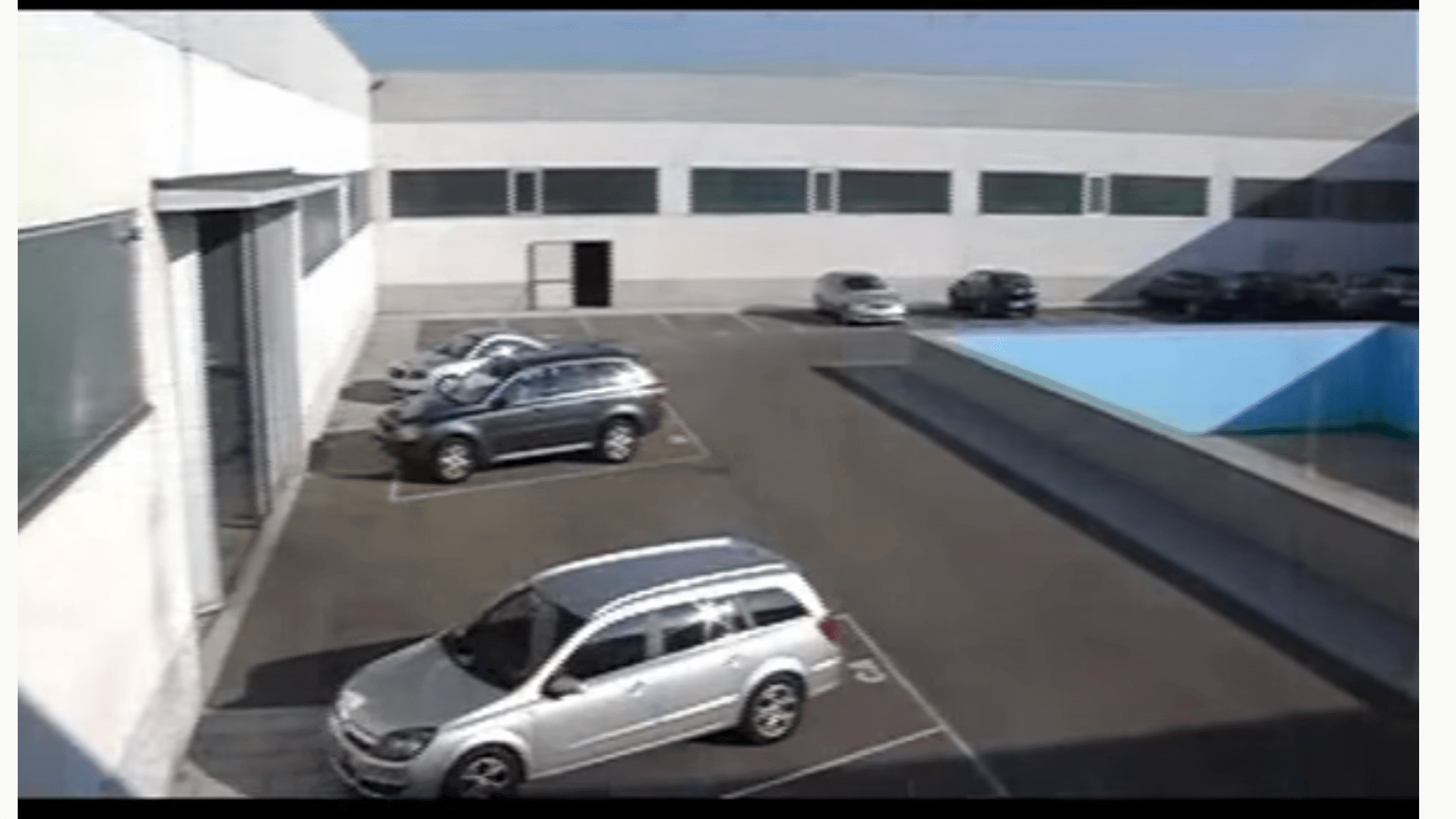}   &   \includegraphics[width=0.22\columnwidth]{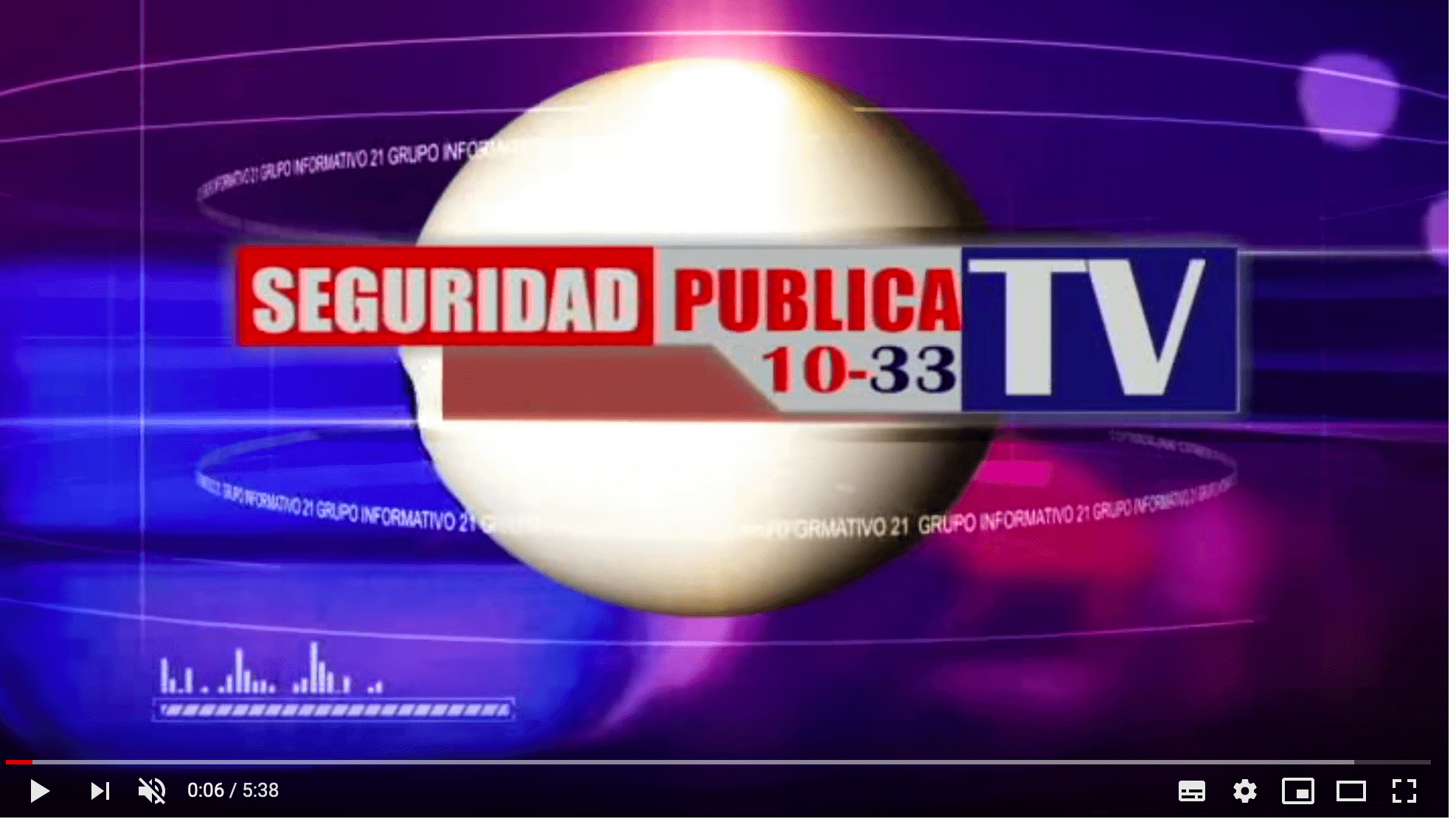}       &   \includegraphics[width=0.22\columnwidth]{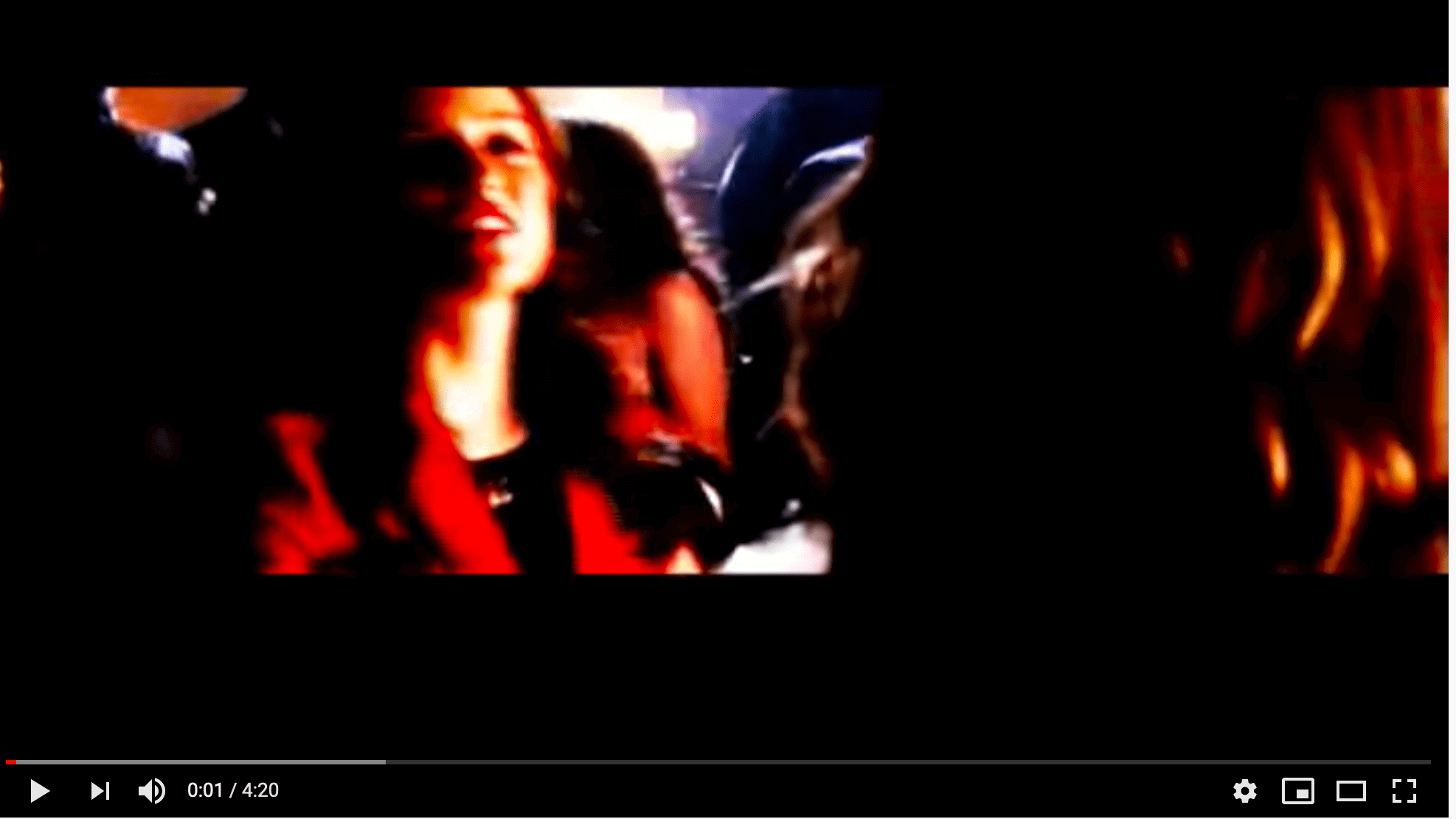}           \\ \hline\hline
Links & bit.ly/2wMvrI2 & bit.ly/39GibU8 & bit.ly/2W3IWOe & bit.ly/2TCo0w9 \\ \hline \hline
      \multirow{4}{*}{GT} 
      		& Trans & Food & Animal & MovieOrTV\\ 
      		& Trans:Land & Food:FI & Animal:Bird  & \\ 
      		& Trans:Land:Car & Food:FI:Dessert & Animal:Bird:Fowl &\\
      		& Trans:Land:Car:Italian & Food:FI:Dairy & Animal:Farm &\\ \cline{1-5}\hline\hline
      \multirow{2}{*}{NV-L} 
      		& Sports:Racing: 0.92 & Trans:Land:Car: 0.89  & MovieOrTV:News: 0.22 &  Music: 0.8\\
      		& Trans:Land:Car: 0.84 & Trans:Land: 0.77 &  Transport: 0.21 &\\  \cline{1-5}\hline\hline
      \multirow{2}{*}{NV-CM-C} 
      		& Sports:Racing: 0.85 & Trans:Land:Car: 0.45 & MovieOrTV: 0.61 & Music: 0.67\\ 
      		& Sports: 0.73 & Trans:Land: 0.43 & MovieOrTV:News: 0.32 & MovieOrTV: 0.33\\  \cline{1-5}
 \hline
\end{tabular}
\vspace{-4mm}
\end{table*}

\section{Related Work}
\label{sec:related}

Multi-modal ML models have been developed for video analysis for a variety of tasks, e.g., video indexing~\cite{snoek2005multimodal}, video search~\cite{francis2019fusion}, emotion detection~\cite{ortega2019multimodal}, virtual reality scene construction~\cite{wang2019virtual}, collaborative learning~\cite{chua2019edubrowser}, 
learning multi-modal representations of contact-rich data types~\cite{lee2019making}, task-agnostic video-linguistic representations~\cite{lu2019vilbert}, explanation~\cite{SelvarajuL2019}, incremental learning~\cite{Li2019,zhang2020rec}, visual dialog~\cite{zhang2019}, and even cyber-security applications like automated categorization of darkweb sites having multimodal (video, text, etc.) content~\cite{ghosh2017}. In this paper, we focus on the multi-modal content recommendation problem as application. 

When we have data in multiple modalities, e.g., audio, video, text, then there are different ways to combine that data. Early fusion concatenates the input embeddings from the different modalities and feeds that combined input embedding to the models --- this is similar to the approach in Ortega et al.~\cite{ortega2019} 
and the Naive Fusion approach in Hori et al.~\cite{Hori2017ICCV}.
Note that the Naive Fusion approach does not allow the attention weights in the fusion to change based on the decoder state. This can be achieved using the Attentional Fusion approach of Hori et al.~\cite{Hori2017ICCV}, which is an interesting example of intermediate fusion. Sparse network fusion~\cite{Narayana_2018_CVPR} is a good example of late fusion, using a temporal aggregation strategy.
In our work, we additionally enforce that when the model is attending to a particular video frame for a task (e.g., object detection) in a particular modality (e.g., image), it will attend to the corresponding frames in the other modalities (e.g., audio, CC-text) with a high probability. Directly enforcing this condition is not appropriate, since the influence of the attention vectors of one modality on the other is not necessarily true --- our approach of enforcing it is a variant of the fine-tuning attention fusion approach of Gu et al.~\cite{Gu2018}. 
The fine-tuning attention fusion is a versatile approach to fuse attention variables for multi-modal modalities --- it uses a dense layer to learn the combination of context but averages the attention vectors. In our approach, we  attenuate/transform the attention vectors of one modality before they are used to influence another modality. Note that this variation makes this formulation similar to the hierarchical attention formulation of Yang et al.~\cite{yang2016}, except in their case there was no cross-modal influence. 
Note that we observe using prior initialization improves model performance, as has been seen too in other models~\cite{ghosh2016,papai2012}. We get the best performance when the attention variable for each modality is used as a prior to the other modality.

In this paper, we have modeled correlation between modalities using a separate sub-network. There are existing papers that model the correlations between audio and video~\cite{korbar2018,sun2019,jiang2011,gillet2007,algur2015} --- our paper uses cross-entropy loss to model the correlation between the audio/video channels and uses hard negative mining to train this sub-network effectively. Another aspect of our work is using DNNs to transform the output of one modality before it's used by another modality --- a similar idea of shifting attention in videos has been used by Long et al.~\cite{long2018}. 

In the context of multi-modal modeling, different types of regularization approaches have been used in the literature, e.g., spatio-temporal regularization~\cite{Kundu2016CVPR}, self-regularization~\cite{lei2019fully}, temporal regularization in different  forms~\cite{Huang2018CVPR,Wang2015,BeckerHinton92,Xiao2019ICCV,MobahiCW09}. Our approach is unique in the sense that the result of one modality is used to regularize another modality --- this is an interesting variation of self-regularization, where one modality regularizes another.

In this paper, we have focused on RNN, Transformer and NetVLAD models. Another model that 
has been applied successfully in jointly modeling video and language data is BERT~\cite{sun2019videobert}--- ensembles of BERT models have been applied to video data as well~\cite{liu2019bert}. In the future, we would like to compare our cross-modal approaches to the relevant BERT models. 

We ran our experiments on a variant of the YouTube-8M dataset~\cite{abu2016youtube}, which we curated internally --- our approaches were able to beat this benchmark. For this work, we have developed a curated hierarchy of micro-genres --- note that such feature-rich hierarchies have been used in object detection and semantic segmentation in other work~\cite{Girshick2014CVPR}.

\section{Conclusions and Future Work}
\label{sec:conclusions}

In this paper, we proposed a novel approach for multi-modal learning with intermediate fusion: cross-modal learning. In this approach, the intermediate result of processing one modality (e.g., the attention vectors in an attention RNN model) is used to influence the other modalities. We further refined this idea to use a correlation network, which predicts when such influence of one modality on the other would be useful. We ran extensive experiments on the YouTube-8M dataset and variants using our hierarchical taxonomy, and showed that models trained using correlation-based cross-modal learning outperform other strong baselines in performance. Our experiments with cross-modal learning show that using the multi-modal correlation prediction as a (concatenated) feature rather than in a gating logic gives higher gains, and certain categories in the data show more gains with cross-modal learning than others. 

In the future, we would like to apply the correlation-based cross-modal formulation to ensemble models of the cross-modal variants. Also, our error analysis of the current approach showed that our correlation tower could be further improved using stronger actual positive and negative examples. We would also like to try other loss formulations (e.g., contrastive loss).

\bibliographystyle{splncs04}
\bibliography{main}

\end{document}